\documentclass{llncs}
\usepackage{makeidx}  
\usepackage{hyperref} 
\usepackage{graphicx} 
\usepackage{booktabs}
\usepackage{xcolor}
\usepackage{tikz}
\usepackage{amssymb}
\usepackage{url}
\usepackage{multirow}
\usepackage[noadjust]{cite}
\usepackage{caption, subcaption, floatrow}

\newcommand{\centered}[1]{\begin{tabular}{l} #1 \end{tabular}}

\usepackage{fancyhdr}
\fancypagestyle{arxivhdr}
{
   \fancyhf{}
   \setlength{\headheight}{50pt} 
\fancyfoot[C]{This paper has been published in: Proceedings of the 16th International Conference on Intelligent Autonomous Systems (IAS 2021)}
\fancyhead[C]{\footnotesize Please cite this paper as:\\
A. Saviolo, M. Bonotto, D. Evangelista, M. Imperoli, J. Lazzaro, E. Menegatti, and A. Pretto,~~``Learning to Segment Human Body Parts with Synthetically Trained Deep Convolutional Networks''. In: Proceedings of the 16th International Conference on Intelligent Autonomous Systems (IAS 2021)}
}

\begin{document}
\frontmatter          
\pagestyle{headings}  
\addtocmark{Hamiltonian Mechanics} 
\mainmatter              
\title{Learning to Segment Human Body Parts with Synthetically Trained Deep Convolutional Networks}
\titlerunning{hand-eye calibration}  

\author{Alessandro Saviolo \inst{1}
		\and 
		Matteo Bonotto \inst{2}
		\and
		Daniele Evangelista \inst{2}
		\and 
		Marco Imperoli \inst{1} 
		\and
		Jacopo Lazzaro \inst{3}
		\and
		Emanuele Menegatti \inst{2}
		\and 
		Alberto Pretto \inst{2}
		}

\authorrunning{Saviolo et al.}

\institute{
    FlexSight Srl, Padova, Italy,\\
    \email{[alessandro.saviolo, marco.imperoli]@flexsight.eu}
\and
	Department of Information Engineering, University of Padova, Padova, Italy,\\
	\email{matteo.bonotto.2@studenti.unipd.it},\\
	\email{[evangelista, emg, alberto.pretto]@dei.unipd.it}
\and
    PlayCast Srl, Padova, Italy,\\
    \email{jacopo@playcast.it}
}

\maketitle
\thispagestyle{arxivhdr}

\begin{abstract}
\label{sec:abstract}
This paper presents a new framework for human body part segmentation based on Deep Convolutional Neural Networks trained using only synthetic data. The proposed approach achieves cutting-edge results without the need of training the models with real annotated data of human body parts.
Our contributions include a data generation pipeline, that exploits a game engine for the creation of the synthetic data used for training the network, and a novel pre-processing module, that combines edge response maps and adaptive histogram equalization to guide the network to learn the shape of the human body parts ensuring robustness to changes in the illumination conditions.
For selecting the best candidate architecture, we perform exhaustive tests on manually annotated images of real human body limbs.
We further compare our method against several high-end commercial segmentation tools on the body parts segmentation task. The results show that our method outperforms the other models by a significant margin.
Finally, we present an ablation study to validate our pre-processing module.
With this paper, we release an implementation of the proposed approach along with the acquired datasets.
\end{abstract}

\section{Introduction}
\label{sec:introduction}

\begin{figure}[t]
    \centering
    \includegraphics[width=\linewidth]{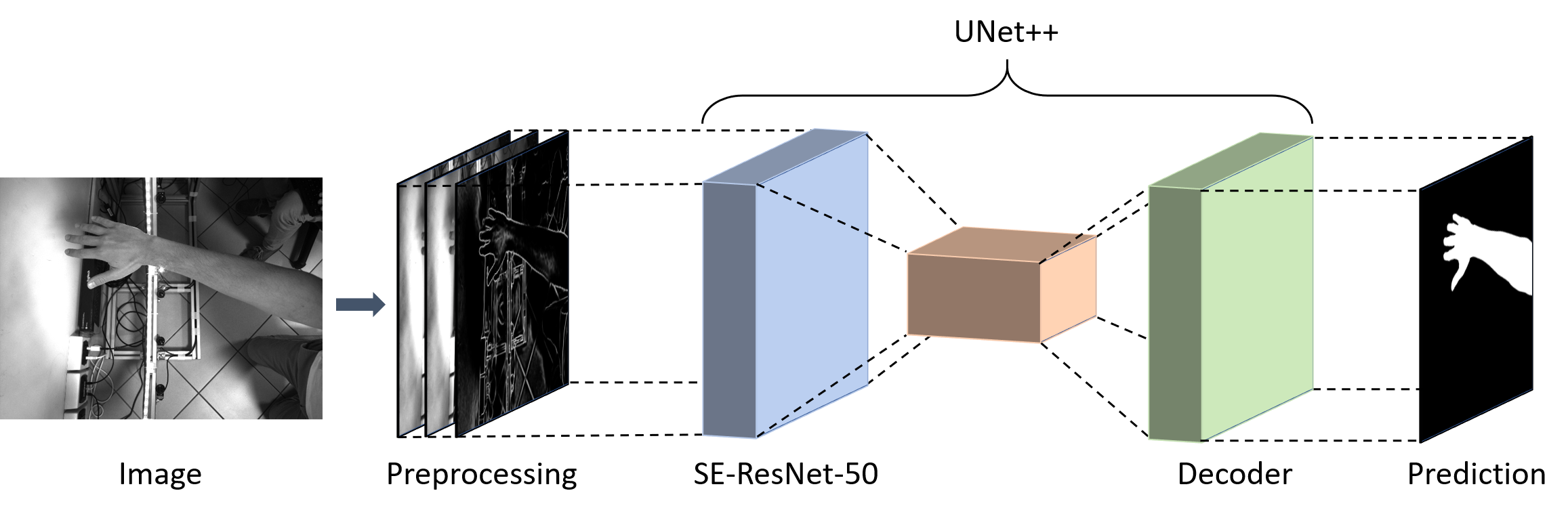}
    \caption{The proposed shape-aware segmentation framework.}
    \vspace{-0.4em}
    \label{fig:segmentation_framework}
\end{figure}

The ability to extract human body parts from the foreground space of images is an essential requirement for many robotics and perception applications such as human-robot interaction \cite{ju2017humanrobot}, surgical robotics and medical images analysis \cite{casas2019medicalapp}, self-perception in mixed reality \cite{gonzalez2020armseg}, and 3D human body reconstruction \cite{smith20193drecon}. A collaborative robot, for example, should be able to localize the arms of the human operator with whom it is collaborating, just as a surgical robot must be able to accurately locate the limb on which it is operating. In egocentric vision systems, hand position recognition is a key capability, for example, to detect which objects come into contact with it. 3D human body reconstruction systems must be able to distinguish which parts of the input images belong to the human body, and which parts must not be considered in the reconstruction.

The body parts localization task can be addressed as a semantic segmentation problem. Semantic segmentation aims to provide a per-pixel classification of the input image. In the specific case of human body segmentation, there are two classes involved: pixels that belong to human limbs and those that do not. Semantic segmentation of naked human body parts can be obtained, for instance, by exploiting skin color detection and segmentation techniques (e.g., \cite{shaik2015colordetection}). Unfortunately, such color-based systems suffer from severe robustness problems due to the variability of skin tone, the variability of light conditions, and the complexity of backgrounds, which can be mistaken for portions of the skin.
Convolutional Neural Networks (CNNs) are convenient and effective methods for addressing the general problem of semantic segmentation. Multiple powerful networks \cite{ronneberger2015unet, zhou2018unetplusplus, chaurasia2018linknet, chen2017deeplabv3, fan2020manet, li2018pan, zhao2017pspnet} have been proposed to obtain superior segmentation results. Unfortunately, these segmentation techniques require thousands of annotated images to be properly trained. Manually labeling the entire dataset is challenging and extremely time-consuming. Furthermore, the quality of the segmentation result (see Sec.~\ref{subsec:performance_evaluation}) in the case of human body parts obtained by most CNN-based systems is fair but certainly not adequate for many of the applications listed above.
Background removal systems can be used for body part localization as well:
unfortunately, such systems either require a rich prior knowledge about the scene to build an accurate background model \cite{lin2020ego2hands, lim2021fgsegnet} or are too general and struggle to adapt to specific tasks.
%
Other techniques, such as background subtraction in a green screen setting \cite{gonzalez2020armseg}, allow to speed up the annotation process. However, they may come at the cost of the unrealistic feature of green color bleeding at the annotation boundaries. 

This paper proposes a new framework for body part segmentation that aims to overcome the limitations listed above. The proposed framework has been specifically trained to obtain highly accurate segmentation masks from images of naked human limbs without the need for any real data annotation since it only relies on synthetically generated data and a custom pre-processing of the network's input.
Our system (see Fig.~\ref{fig:segmentation_framework}) is based on a pipeline that includes a state-of-the-art semantic segmentation architecture \cite{zhou2018unetplusplus} and a custom pre-processing module that feeds the segmentation network with a specific input tensor built by concatenating the edge response map and two equalized versions of the original input image. The pre-processing module is designed to polarize the network to focus on the shape of the objects of interest (human limbs in our case) rather than the skin tone (intensity or color) or its characteristic texture. This simple modification greatly increases the level of segmentation accuracy compared to the same network fed with raw input images. Moreover, to emphasize skin tone independence, we performed all the experiments considering gray-level images. Our system also does not require any manually annotated data for training. The data are automatically generated by rendering, with a game engine, synthetic photo-realistic views of human limbs: such images are provided with ground truth segmentation masks (i.e., the annotations) at no additional cost. The geometric properties of human limbs (e.g., scale, deformation) and the scene configuration (e.g., illumination, camera positions) are defined by a set of parameters. By randomly choosing a combination of these parameters within predefined ranges, our data generation pipeline can generate multiple unique synthetic images and their corresponding annotations.

We report an exhaustive performance evaluation, focusing on the arm segmentation use case. Experiments have been performed on a custom-built, manually labeled dataset that includes several views of real human arms. Our experiments aim both to validate the architectural choices and to show that our system overcomes by a large margin several state-of-the-art segmentation architectures in the specific human limb segmentation task. Furthermore, we compare our method against several cutting-edge commercial foreground segmentation networks on the body parts segmentation task, and demonstrate that the considered networks are too generic to produce accurate segmentation results.

An open-source implementation of the proposed system along with the generated dataset is made publicly available with this paper at:\\ \url{https://github.com/AlessandroSaviolo/HBPSegmentation}

\section{Related Work}
\label{sec:related_work}

\noindent \textbf{Segmentation models.} As for many other computer vision tasks, deep convolutional networks have been proven highly effective to perform semantic segmentation. Multiple powerful networks have been proposed to obtain superior segmentation results. Pioneering work \cite{ronneberger2015unet} proposed the UNet network, an encoder-decoder architecture consisting of a contracting path to capture context and a symmetric expanding path that enables precise localization. Even though the network was developed for biomedical image segmentation, it generalized well to other domains, such as urban scenarios \cite{tao2020hierarchical} and indoor environments \cite{pigny2020indoorunetapp}. Due to this, numerous variations of the UNet network have also been proposed. Among all, the UNet++ \cite{zhou2018unetplusplus} demonstrated promising results. Such a network modifies UNet by connecting the encoder to the decoder through a series of nested, dense skip pathways. Other more recent pioneering work include DeepLabV3 \cite{chen2017deeplabv3}, PSPnet \cite{zhao2017pspnet}, PAN \cite{li2018pan}, LinkNet \cite{chaurasia2018linknet}, and MAnet \cite{fan2020manet}. In this work, we report an extensive evaluation of these networks for the task of human body part segmentation where the limb is in the foreground of the image.

\noindent \textbf{Synthetic data generation.} Training segmentation networks requires a large amount of annotated data, which can either be collected from the real world or generated synthetically. Collecting data from the real world involves either annotating manually each image or using a background removal technique in a green screen setting. Since the former is too time-consuming, many researchers have focused on the latter, by generating various datasets for multiple tasks, such as hand segmentation \cite{li2013handsegm} and activity recognition \cite{li2015activityrec}. However, these datasets lack variety and accuracy (due to the unrealistic feature of green color bleeding at the annotation boundaries) for sufficiently training segmentation models for real-world applications. Recent work in computer vision looks at using synthetic data to train segmentation networks, for example in \cite{dicicco2017freecitation} authors addressed the semantic segmentation problem in agricultural scenarios by proposing an automatic, model-based dataset generation procedure that generates large synthetic training datasets by rendering a large amount of photo-realistic views of an artificial agricultural environment with a modern 3D graphic engine. Contrary to collecting real data, generating synthetic images allows to introduce a tremendous amount of data variability and comes at minimum cost. However, despite the increasing realism of such synthetic data, there remain perceptual differences between real and synthetic images. To address this so-called \emph{reality gap}, multiple techniques have been proposed, such as creating photo-realistic renders \cite{johnson2016photorealisticdata}, adopting domain randomization techniques \cite{tobin2017domainrandomization}, and learning only the decoder layers of networks while freezing the encoder pre-trained on real data \cite{hinterstoisser2017freezingencoder}. In this work, we apply many of these techniques, but also we propose a novel shape-aware pre-processing module to bridge the reality gap.

\textbf{Foreground segmentation.} Foreground segmentation, also known as background subtraction, is one of the most fundamental and critical tasks in computer vision. The aim of this task is to differentiate between pixels belonging to foreground and background.
In recent years, foreground segmentation has attracted extensive attention. As a consequence, a host of advanced segmentation models have been proposed.
Most conventional approaches rely on building a background model of the scene \cite{lin2020ego2hands, lim2021fgsegnet}. However, to build an accurate background model one must have rich prior knowledge about the environment, which is not always available.
Following data-driven approaches, other methods that rely only on the information provided by the single input image have been proposed \cite{removebg2021website, pixlr2021website, slazzer2021website, photoshop2021website, removalai2021website, photoscissors2021website}. These methods have stunning results, offering high-quality segmentation even on complex regions of the images, e.g. feathery hair. But, the generality of such systems comes at the cost of being poorly adaptable to specific tasks, such as human body parts segmentation.
In this work, we propose a novel method for accurate and robust human body foreground segmentation which relies only on the information provided by the single input image.

\section{Body-Part Synthetic Data Generation}
\label{sec:data_pipeline}

\begin{figure}[t]
    \centering
    \includegraphics[width=0.85\linewidth]{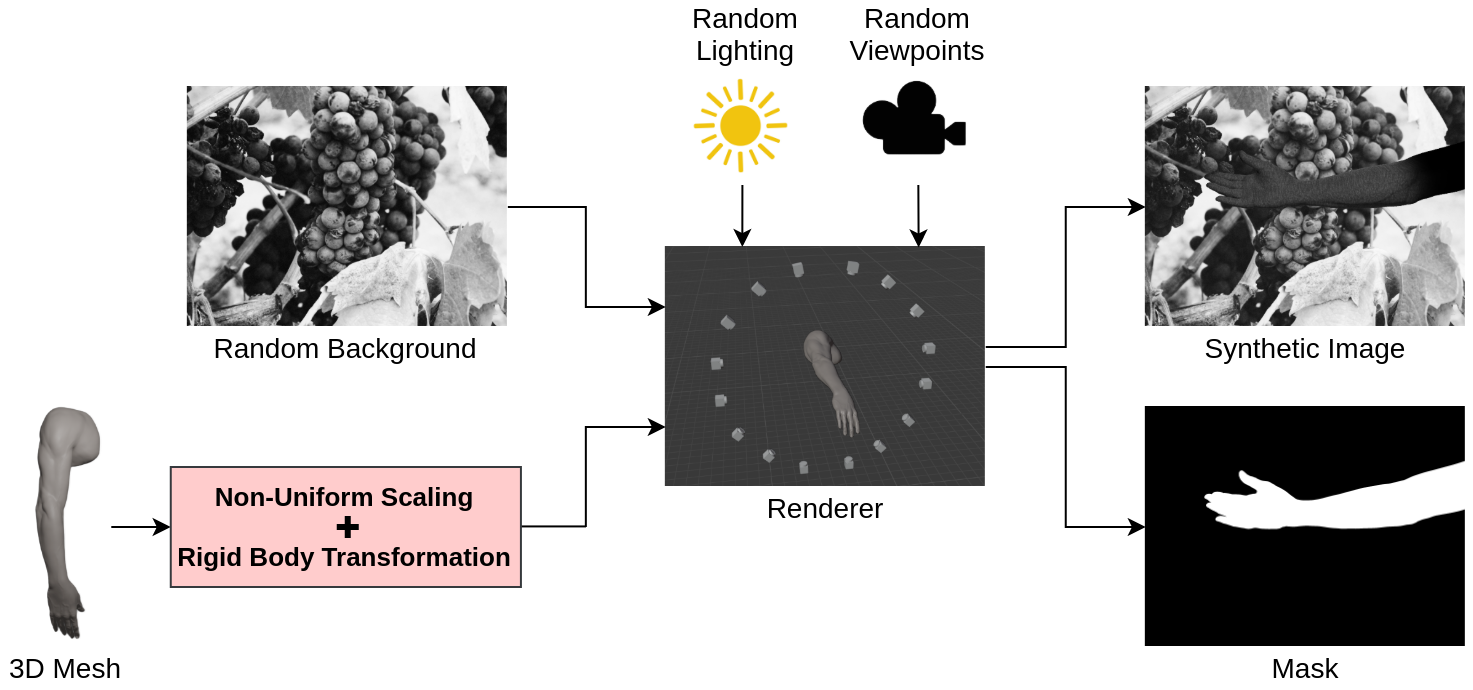}
    \caption{The proposed synthetic data generation pipeline.
    }
    \label{fig:pipeline}
\end{figure}

The front-end of our framework is a synthetic data generation pipeline that enables to produce efficiently and effectively synthetic annotated images for training and evaluating segmentation networks. The input of our pipeline (Fig.~\ref{fig:pipeline}) is a set of 3D meshes of human limbs. Each mesh is positioned with a random location and orientation in an empty scene. A random number of different types of lighting sources are inserted at random locations in the scenes. Each scene is framed by multiple virtual cameras positioned in random viewpoints. For every acquired view, we sample from a uniform distribution the location and orientation of the framed limb, and the positions and intensity of the light sources. Similarly to \cite{tremblay2018simtoreal}, we also apply to the limbs a non-uniform scaling transformation, with the directional scaling factors $s_x,s_y,s_z$ sampled from a uniform distribution. Moreover, we apply to each limb a random texture defined by multiple shades of gray and irregular dark dots representing skin flaws (e.g., moles). Finally, each image is stored with the corresponding annotation, which is computed by rendering the limb on a black background and then applying a binary threshold to the image.

The synthetic data generation pipeline pursues two objectives: i) Maximize the variability of the data, since with enough variability in the simulated environment, the real-world may appear as just another variation of the synthetic scene; ii) Emphasize the shape of the mesh in the images, to guide the networks at training time to focus on relevant features, ignoring useless details that negatively affect the quality of the predictions. The first objective is achieved by defining the geometric properties of human limbs and the scene configuration by a set of parameters, and then randomly choosing a combination of these parameters within predefined ranges. The second objective is achieved by adopting a domain randomization strategy. By applying random backgrounds to the images, networks are forced to focus at training time on the shape of the object of interest and ignore other objects in the scene (i.e., networks better learn to focus on the relevant features, rather than the skin tone or its characteristic texture).




\section{Shape-Aware Segmentation}
\label{sec:network}

In this section, we present our segmentation framework based on a state-of-the-art segmentation network and a custom pre-processing module. Such a module feeds the segmentation network with a specific input tensor built by concatenating the edge response map and two equalized versions of the input image.

As introduced, semantic segmentation frameworks based on CNNs provide state-of-the-art results and currently represent the gold standard in several settings, among others medical images \cite{zhou2018unetplusplus}, urban environments \cite{tao2020hierarchical}, and agricultural scenarios \cite{pretto2020ram}. 
Unfortunately, the direct application of CNNs segmentation models to the human body segmentation task, trained on synthetic data (see Sec.~\ref{sec:data_pipeline} and Sec.~\ref{subsec:exp_setup}), did not lead to satisfactory results (see Sec.~\ref{subsec:performance_evaluation}).
One of the major problems of such systems is the tendency to generate artifacts in the segmentation masks, with shapes that differ from human limb shapes by a large margin (e.g., see Tab.~\ref{tab:preprocess_ablation_qual}, right column). We also found that the robustness of such systems isn't always adequate against the variability of light conditions.

To induce the segmentation networks to minimize the generation of artifacts in the segmentation masks, we propose to augment the input of the segmentation networks with an \emph{edge map} extracted from the image. Such augmentation aims to induce the network to focus on the shape of the human limbs to be segmented, rather than the skin tone (intensity or color) or its characteristic texture. Specifically, we extract from the image a map that for each pixel encodes the edge responses, that is a real value ranging from $0$ (no edge) to $1$ (maximum probability that the pixel is an edge). We call this map \emph{edge response map}, to distinguish it from a binarized edge map.

To make the segmentation networks more robust against the variability of light conditions, we propose to replace the input image with its equalized version. By equalizing the input image, we improve the contrast and therefore the invariance under different light conditions.
The edge response map and the equalized image are then concatenated and fed to the segmentation network. This simple modification enables a considerable increase in the level of segmentation accuracy compared to the same segmentation network fed with raw input images.

Our pre-processing module (Fig.~\ref{fig:segmentation_framework}) relies on the Holistically-Nested Edge Detection (HED) \cite{xie2015hed} method to extract the edge response map. HED is a CNN-based edge detection system that enables cutting-edge performance by combining multi-scale and multi-level visual responses: this allows it to be independent of the scale of the objects whose border is to be extracted. Histogram equalization is obtained by exploiting the Contrast Limited Adaptive Histogram Equalization (CLAHE) method \cite{zuiderveld1994clahe}. CLAHE is an adaptive method that computes several local histograms, each corresponding to a specific image portion, 
so it is well suited for improving the contrast also in images that present different lighting conditions inside them. To avoid amplifying noise, CLAHE applies contrast limiting: if any histogram bin is greater than a contrast threshold, the corresponding pixels are distributed uniformly to other bins before applying histogram equalization. To be more independent of this threshold, in our system we apply two different contrast thresholds and concatenate the two resulting images with the edge response map extracted with HED (Fig.~\ref{fig:segmentation_framework}).
The generated tensor is finally given as input to the UNet++ segmentation network, which from our tests resulted in the best segmentation network when applied to human body parts (see Sec.~\ref{subsec:performance_evaluation}).
To validate the composition of our input tensor, we performed an exhaustive ablation study (see Sec.~\ref{subsec:ablation_study}) showing superior results when comparing with alternative combinations.

\section{Experiments}

We design our evaluation procedure to address the following questions: What is the best synthetically-trained state-of-the-art network for the task of human body part segmentation? How can we improve the network performances to generate more accurate results? How does the network perform in comparison with the cutting-edge commercial foreground segmentation tools? Finally, we validate our design choices with ablation studies.

\subsection{Experimental setup} \label{subsec:exp_setup}

In this section, we describe the simulated and real environments that we designed to evaluate the proposed framework. Moreover, we present the datasets used in our experiments and provide details about the considered segmentation networks and the adopted training and evaluation procedures.\\

\noindent \textbf{Simulation Environment.} In our experimental setup, the choice of the simulator takes a central role. The simulator must guarantee a high degree of flexibility when designing the scene, a deep level of customization of the imported objects, and a user-friendly interface. To achieve this, we exploited the Unity3D\footnote{\url{https://unity.com/}} game engine as the basic software infrastructure for designing our simulated scene.

The simulated scene has been designed following the structure of the Renderer depicted in Fig.~\ref{fig:pipeline}. The key components of the simulated scene are the cameras, the lighting sources, and the 3D mesh of the human limb. In particular, we designed a ring of $16$ uniformly distributed cameras and placed $4$ additional lighting sources to illuminate the scene. The set of cameras has been used to acquire $360$ degrees views of the 3D models with a one-shot acquisition setting. The whole synthetic dataset has been collected by iteratively reproducing the aforementioned acquisition setup and applying a randomized set of parameters at each iteration. The parameter set has the following characteristics:
\begin{itemize}
    \item \textbf{3D Mesh Selection:} $14$ high-quality 3D meshes of human limbs have been selected and downloaded from \emph{TurboSquid}\footnote{\url{https://turbosquid.com/}}, each one representative of certain characteristics (e.g., size, shape). We randomly utilized one of them at every iteration.
    \item \textbf{3D Mesh Transformation:} scale in the range of $[0,5] \%$ on each axis; pose translation in the range of $[0, 0.01]$ meters on each axis; pose rotation in the range of $[0, 25]$ degrees on each axis.
    \item \textbf{Scene Lighting Conditions:} light source intensity has been randomly set for each light source in the scene.
    \item \textbf{Background Selection:} for each image captured, we compose the rendered model with a random natural background image drawn from a set of $1000$ manually downloaded images.
\end{itemize}

\begin{figure}[t]
    \centering
    \subfloat[\centering High]{{\includegraphics[width=10.5em]{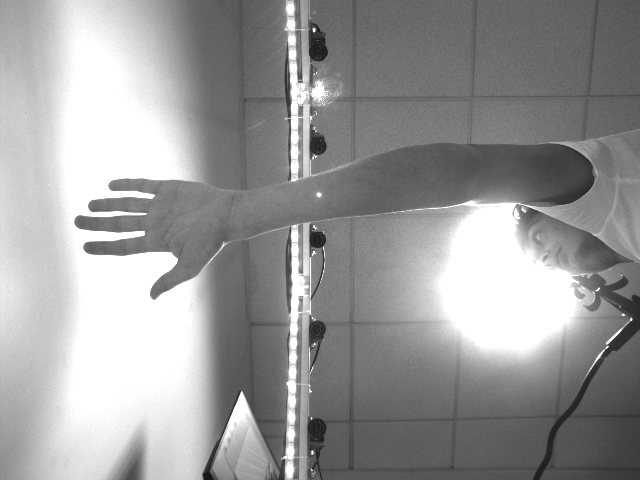}}} \hfill
    \subfloat[\centering Medium]{{\includegraphics[width=10.5em]{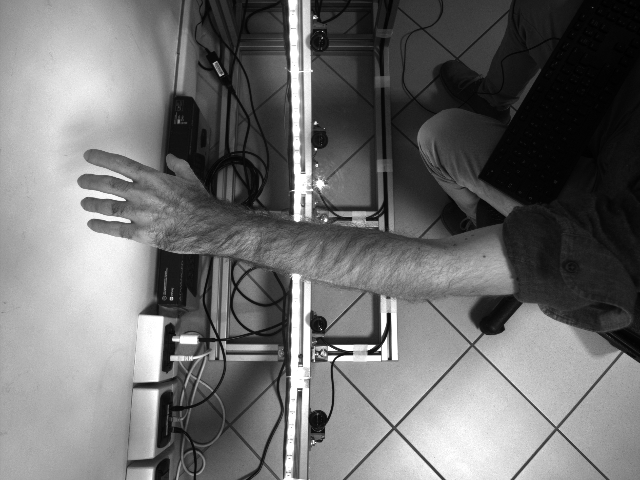}}} \hfill
    \subfloat[\centering Low]{{\includegraphics[width=10.5em]{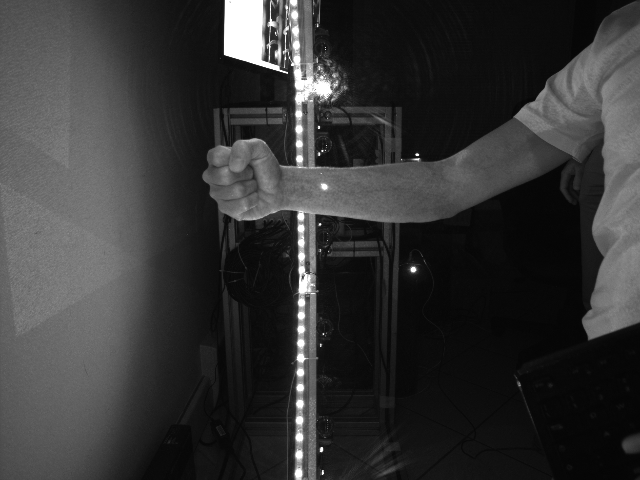}}}
    \vspace{-0.59em}
    \caption{Examples of test instances.}
    \label{fig:test_examples}
\end{figure}

\noindent \textbf{Synthetic Dataset.} Using the data generation pipeline described in Sec.~\ref{sec:data_pipeline} and the simulated environment described above, we produced a synthetic dataset consisting of $11200$ labeled images of size $640 \times 480$. Moreover, to emphasize skin tone independence, all the synthetic images are in grayscale. The synthetic dataset has been used both for training and evaluating the models, by splitting it into $80\%$, $20\%$ training, and validation sets respectively.

To further increase the synthetic data variability, we apply data augmentation by randomly applying the following transformations to the input images before training: left-right or top-bottom flip, change of brightness, gamma or contrast, application of blur, motion blur, or sharpening filters.\\

\noindent \textbf{Real Dataset.} To test the effectiveness and robustness of the trained models to complex backgrounds and different lighting conditions, we have reproduced the ring of cameras of the simulated scene described above and placed the ring in an office room with multiple objects, such as cables, monitors, and chairs, in the scene. The office room presents multiple challenges, such as the presence of numerous artificial sources of illumination. Moreover, we placed the ring vertically, so that the background captured by each camera is different.

Using this environment, we collected three datasets diversified by specific lighting conditions:
\begin{itemize}
    \item \textbf{High}: set of $48$ images collected with both natural and artificial light. The natural light brights the scene from two large windows, and the artificial light is mainly emitted by multiple sources attached to the ceiling of the office room. The mean and standard deviation of the pixel intensity of the images are, respectively, $234.92$ and $117.88$.
    \item \textbf{Medium}: set of $32$ images collected with only artificial light. The light is mainly emitted by multiple sources attached to the ceiling of the office room. The mean and standard deviation of the pixel intensity of the images are, respectively, $80.00$ and $44.47$.
    \item \textbf{Low}: set of $16$ images collected without any strong source of light. The mean and standard deviation of the pixel intensity of the images are, respectively, $37.49$ and $30.05$.
\end{itemize}
Examples of test instances are illustrated in Fig.~\ref{fig:test_examples}. After the collection, the data have been manually annotated for test purposes.\\

\noindent \textbf{Networks.} In this work, we considered encoder-decoder networks due to their proven robustness and versatility for the task of semantic segmentation \cite{zhao2020encdecsota}. In these networks, the encoder generates a high-dimensional feature vector from the input image, and the decoder produces the output segmentation mask from the feature vector. Multiple encoder-decoder networks have been proposed for segmenting images in various computer vision tasks. In particular, as already introduced in Sec.~\ref{sec:network}, the proposed body limb segmentation pipeline involves the usage of the UNet++ network with a SE-ResNet-50 \cite{hu2018senet} encoder. This network outperforms the original UNet by considering densely connected convolutional layers introduced with DenseNet.

The proposed network has been selected after a first performance evaluation step that will be presented in Sec.~\ref{subsec:performance_evaluation}. The evaluation involved other convolutional networks and encoders used for segmenting images in various computer vision tasks. Specifically, besides of UNet and UNet++, the other considered networks were: LinkNet, DeepLabV3, MAnet, FPN, PAN, PSPnet. Moreover, we considered multiple encoders for each network: DPN-68 \cite{chen2017dpn}, EfficientNet-B4 \cite{tan2019efficientnet}, MobileNet-V2 \cite{sandler2018mobilenet}, VGG-19 \cite{simonyan2014vgg19}, Xception \cite{chollet2016xception}, ResNet-34, ResNet-50 \cite{he2016resnet}.

Each considered network was initialized with weights pre-trained on the Imagenet dataset \cite{deng2009imagenet}, and then retrained on the proposed synthetic dataset for $66$K iterations using a batch size of $4$. During training, we used the Adam optimizer and set both the learning rate and epsilon to $0.0001$. After training, each network was evaluated on the High, Medium, and Low test sets. To compare the performances, we adopted the intersection over union (IoU) metric due to its ease of interpretability:

\begin{equation}
  \textit{IoU} = \frac{T_{p} }{T_{p} + F_{p} + F_{n}}
 \label{eq:iou}
\end{equation}

where $T_{p}$, $F_{p}$, and $F_{n}$ are the number of true positives, false positives, and false negatives, respectively, for the class representing pixels that belong to human body parts.

All the training routines were performed on an NVIDIA Titan X GPU with 24GB of dedicated memory, while the inference operations were executed on a GeForce RTX 2060 GPU with 6GB of dedicated memory.

\subsection{Performance Evaluation}
\label{subsec:performance_evaluation}

\begingroup
\setlength{\tabcolsep}{4pt}
\begin{table}[t]
    \centering
    \captionsetup[subtable]{position=below}
    \captionsetup[table]{position=top}
    \begin{subtable}{0.49\linewidth}
        \centering
        \begin{tabular}{c c c c c}
            \toprule
            Network & High & Medium & Low\\
            \midrule
            DeepLabV3 & \textbf{0.762} & 0.535 & 0.185 \\
            LinkNet & 0.350 & 0.315 & 0.056 \\
            FPN & 0.677 & 0.402 & 0.041 \\
            MAnet & 0.753 & \textbf{0.655} & \textbf{0.269} \\
            PAN & 0.612 & 0.378 & 0.051 \\
            PSPNet & 0.488 & 0.401 & 0.038 \\
            UNet & 0.677 & 0.602 & 0.241 \\
            UNet++ & \textbf{0.796} & \textbf{0.708} & \textbf{0.397} \\
            \bottomrule
        \end{tabular}
        \caption{}
        \label{tab:architecture_comparison}
    \end{subtable}
    \begin{subtable}{0.49\linewidth}
        \centering
        \begin{tabular}{c c c c c}
            \toprule
            Encoder & High & Medium & Low\\
            \midrule
            DPN-68 & 0.790 & 0.590 & 0.215\\ 
            EfficientNet & \textbf{0.860} & \textbf{0.735} & 0.387\\ 
            MobileNet & 0.749 & 0.647 & 0.412\\ 
            ResNet-34 & 0.796 & 0.708 & 0.397\\ 
            ResNet-50 & 0.761 & 0.502 & 0.281\\ 
            SE-ResNet-50 & 0.808 & \textbf{0.734} & \textbf{0.499}\\ 
            VGG-19 & 0.765 & 0.480 & 0.217\\ 
            Xception & \textbf{0.854} & 0.586 & \textbf{0.466}\\ 
            \bottomrule
        \end{tabular}
        \caption{}
        \label{tab:encoder_comparison}
    \end{subtable}
    \caption{Networks (a) and encoders (b) performance analysis. In (a) multiple networks are evaluated using ResNet-34 encoder. In (b) UNet++ network is evaluated with different encoders. For each test set, the two best scores are reported in \textbf{bold}. Segmentation scores are reported in terms of the intersection over union.}
    \label{tab:performance_evaluation}
\end{table}
\endgroup

In this section, we extensively evaluate the choices made during the design process of our system, by training multiple state-of-the-art networks on the proposed generated synthetic dataset and then evaluating their performances on the High, Medium, and Low test sets. 

The first design choice regards the selection of the network. Tab.~\ref{tab:architecture_comparison} summarizes the experimental results obtained by training each network on synthetic data and evaluating it on the three different real datasets. To make the comparison fair, we fixed ResNet-34 as the encoder for all the segmentation models. The results show that the UNet++ network is capable of better learning to calculate rich and proper features that are important for human limb segmentation. This is especially demonstrated by the higher score achieved by the network on the Low test set, where most of the models proved very poor performances.

Even though UNet++ with ResNet-34 as encoder gave encouraging results in the High and Medium test sets and achieved the best performances in the Low test set, the IoU score is still insufficient. Therefore, we performed an evaluation analysis of the encoder choice. Tab.~\ref{tab:encoder_comparison} summarizes the experimental results obtained by training a UNet++ network with different encoders on synthetic data and evaluating it on the three different real datasets. The results show that EfficientNet, SE-ResNet-50, and Xception are the top choices for the encoder of UNet++. Specifically, SE-ResNet-50 demonstrates to be the best encoder due to the higher score in the two more complex test sets, namely Medium and Low.

\begingroup
\setlength{\tabcolsep}{9pt}
\begin{table}[t]
    \centering
    \begin{tabular}{c c c c c c}
        \toprule
        Network & Encoder & Pre-process & High & Medium & Low\\
        \midrule
        UNet++ & SE-ResNet-50 & & 0.808 & 0.734 & 0.499 \\
        UNet++ & SE-ResNet-50 & \checkmark & \textbf{0.946} & \textbf{0.959} & \textbf{0.892} \\
        \bottomrule
    \end{tabular}
    \vspace{1em}
    \caption{IoU results of the proposed approach with and without pre-processing module. For each test set, the best score is reported in \textbf{bold}.}
    \label{tab:preprocess_ablation_quant}
\end{table}
\endgroup

\subsection{Results} \label{subsec:results}

In this section, we analyze and compare the results generated by our system and several cutting-edge commercial software widely known for their accurate performances on the foreground segmentation task. Furthermore, we demonstrate the importance of the shape-aware pre-processing module.

Tab.~\ref{tab:our_vs_commercial} shows the segmentation masks produced by our method and several commercial software available on the Web, namely RemoveBG \cite{removebg2021website}, Pixlr \cite{pixlr2021website}, Slazzer \cite{slazzer2021website}, Photoshop \cite{photoshop2021website}, RemovalAI \cite{removalai2021website}, and PhotoScissors \cite{photoscissors2021website}. The results show that the commercial tools for foreground segmentation are capable of accurately extracting the foreground objects from single input images without relying on prior knowledge about the scene. However, these methods are too generic and can not be used to focus only on a specific object in the foreground space, such as human body parts. On the contrary, our method is specifically tailored to our task and learns to segment human body parts accurately and robustly. By leveraging the shape information of the human limbs, the proposed method is robust to the variability of skin tone, the variability of light conditions, and the complexity of backgrounds.

The direct application of deep convolutional networks to the human body segmentation task, trained on synthetic data (see Sec.~\ref{sec:data_pipeline} and Sec.~\ref{subsec:exp_setup}), did not lead to satisfactory results (see Sec.~\ref{subsec:performance_evaluation}). Indeed, we proposed to pre-process the input data by using a combination of Holistically-Nested Edge Detection (HED) and Contrast Limited Adaptive Histogram Equalization (CLAHE) with two different contrast limiting thresholds (named CLAHE2 and CLAHE4 in following sections).

In Tab.~\ref{tab:preprocess_ablation_quant} and Tab.~\ref{tab:preprocess_ablation_qual}, we give quantitative and qualitative results produced by the network selected in Sec.~\ref{subsec:performance_evaluation} with and without the pre-processing module. Quantitative results show that the pre-processing step boosts the network performance up to $45 \%$. This is further illustrated by the qualitative results, where it is visually noticeable how the network focuses more on the shape of the human limb when the pre-processing module is used.

Even though the proposed system was fully trained with synthetic images, it achieves high accuracy in predicting the segmentation masks. Moreover, it demonstrated to be very robust to changes in lighting conditions and background configurations. Results show the power of our approach in bridging the gap between the simulated and real-world domains.

\begingroup
\begin{table}
    \centering
    \begin{tabular}{c c c c c}
        \toprule
        \rotatebox[origin=c]{90}{Image} &
        \centered{\includegraphics[width=8em]{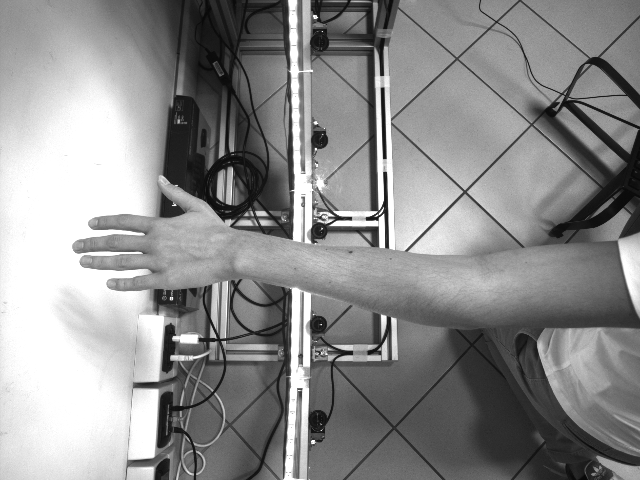}} &
        \centered{\includegraphics[width=8em]{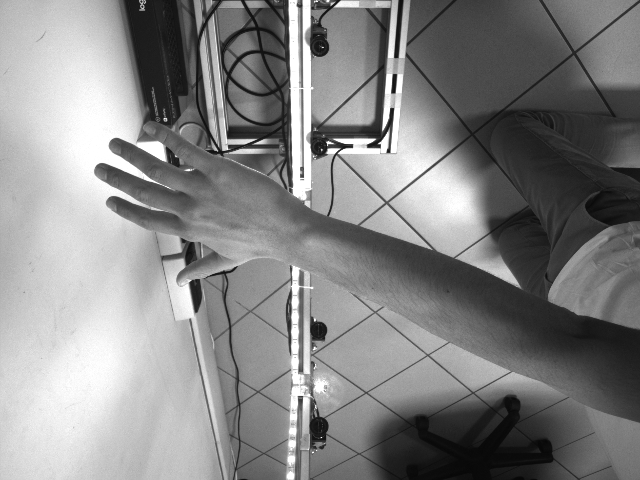}} &
        \centered{\includegraphics[width=8em]{images/dark_light/imgs/img_15.png}} &
        \centered{\includegraphics[width=8em]{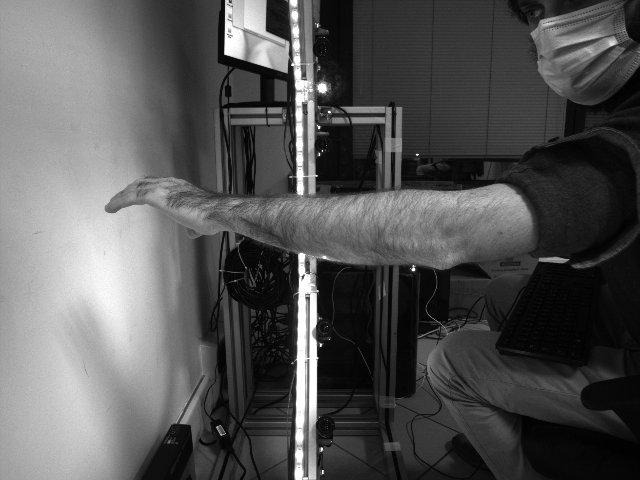}} \\
        \rotatebox[origin=c]{90}{Ground Truth} &
        \centered{\includegraphics[width=8em]{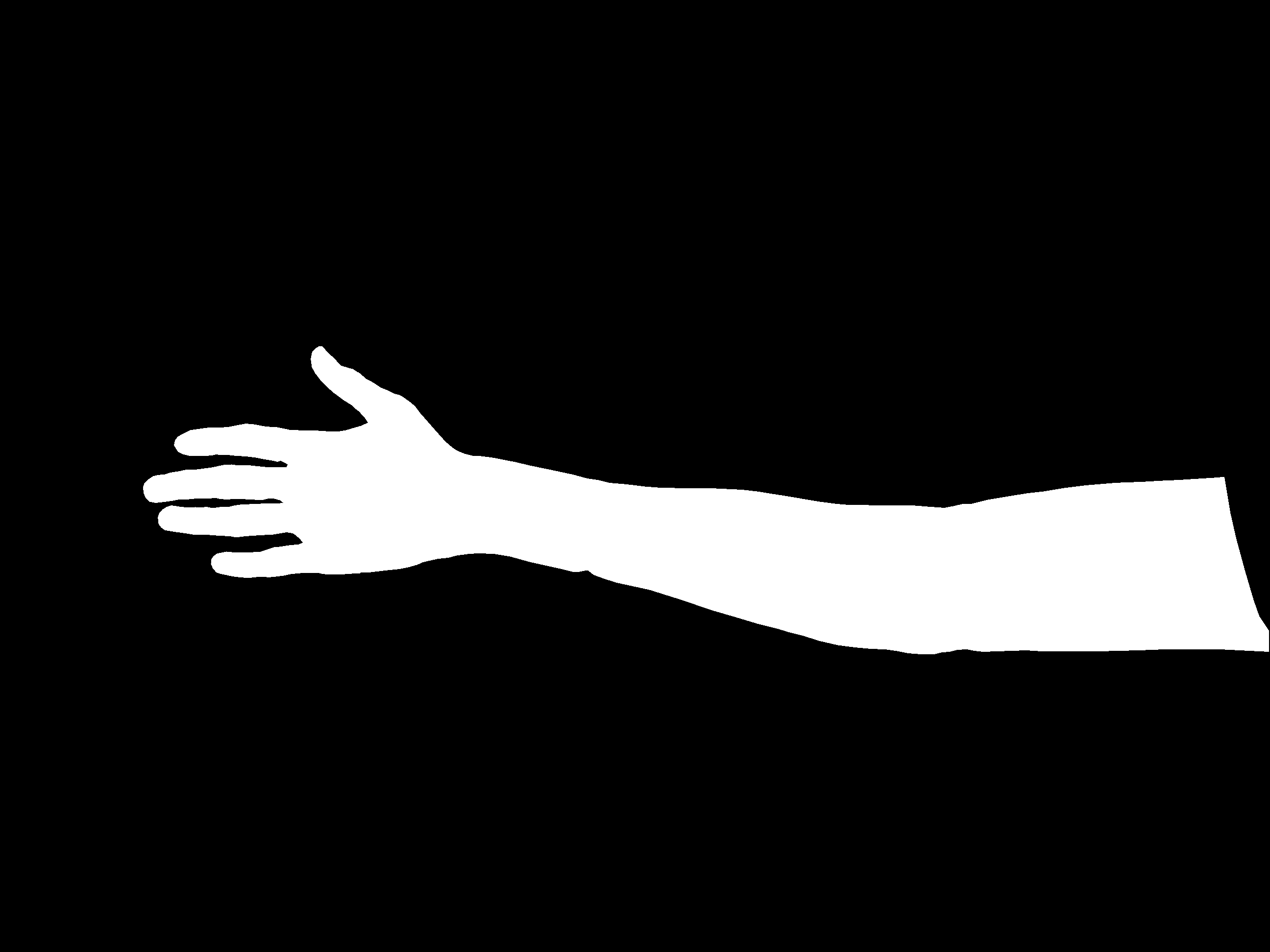}} &
        \centered{\includegraphics[width=8em]{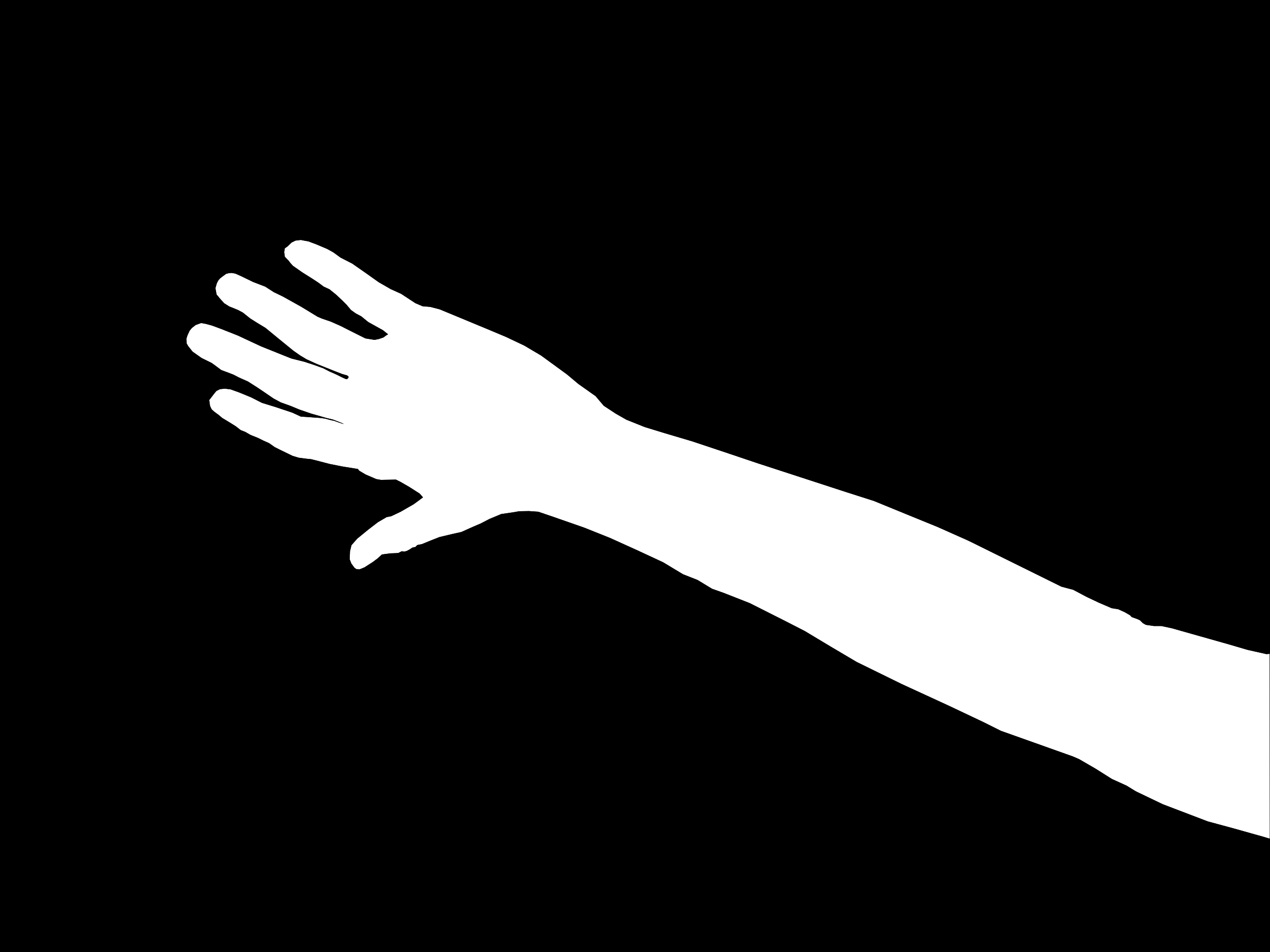}} &
        \centered{\includegraphics[width=8em]{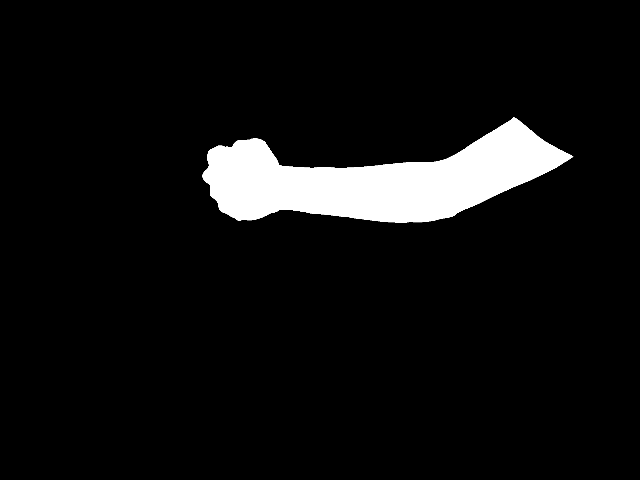}} &
        \centered{\includegraphics[width=8em]{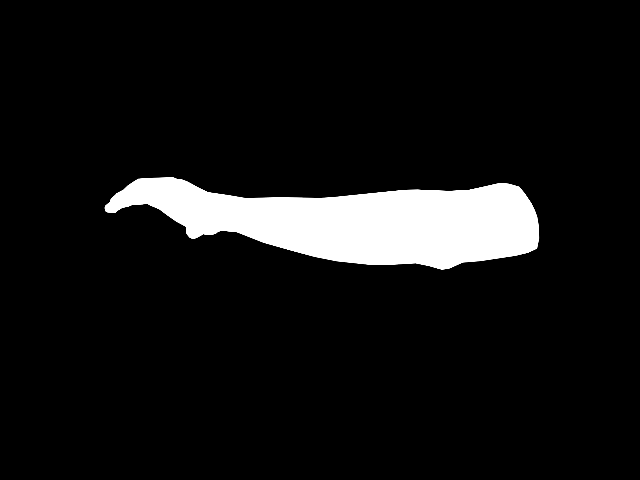}} \\
        \rotatebox[origin=c]{90}{Our Method} &
        \centered{\includegraphics[width=8em]{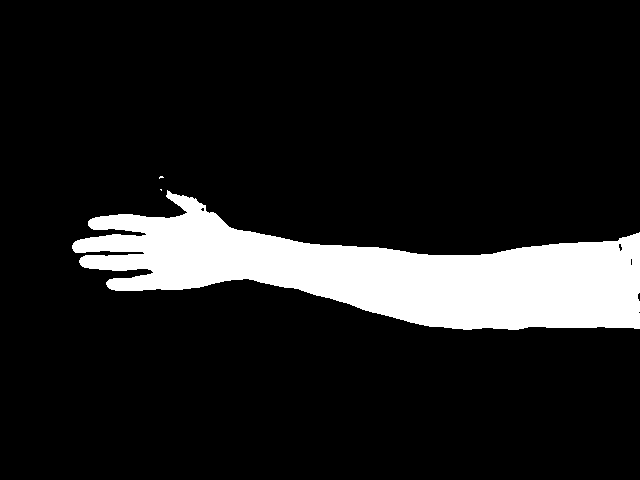}} &
        \centered{\includegraphics[width=8em]{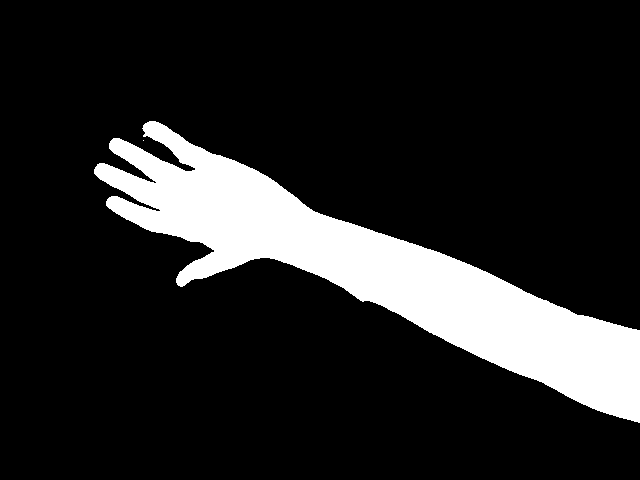}} &
        \centered{\includegraphics[width=8em]{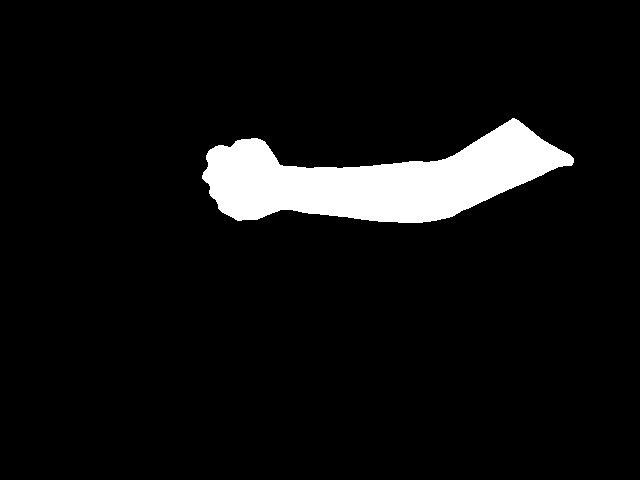}} &
        \centered{\includegraphics[width=8em]{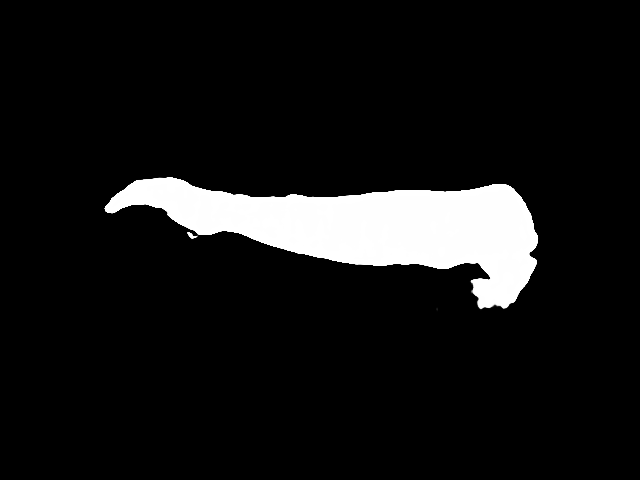}} \\
        \rotatebox[origin=c]{90}{\cite{removebg2021website}} &
        \centered{\includegraphics[width=8em]{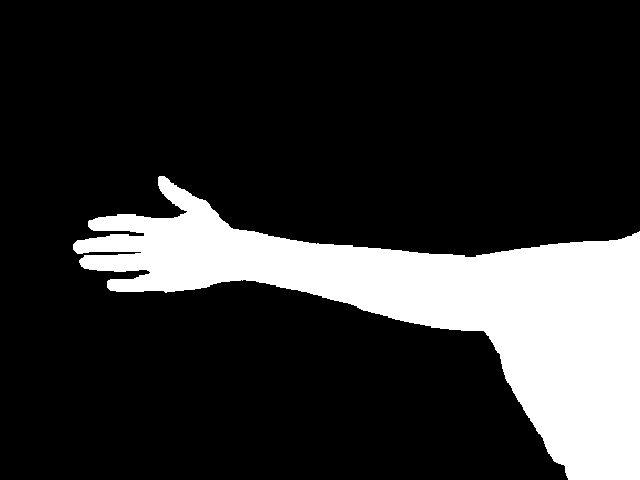}} &
        \centered{\includegraphics[width=8em]{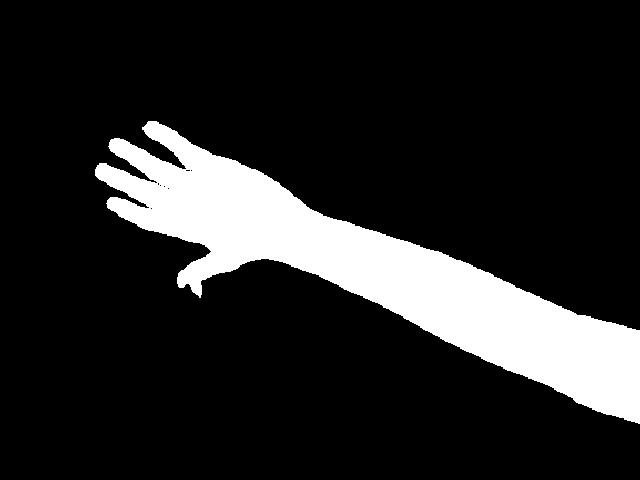}} &
        \centered{\includegraphics[width=8em]{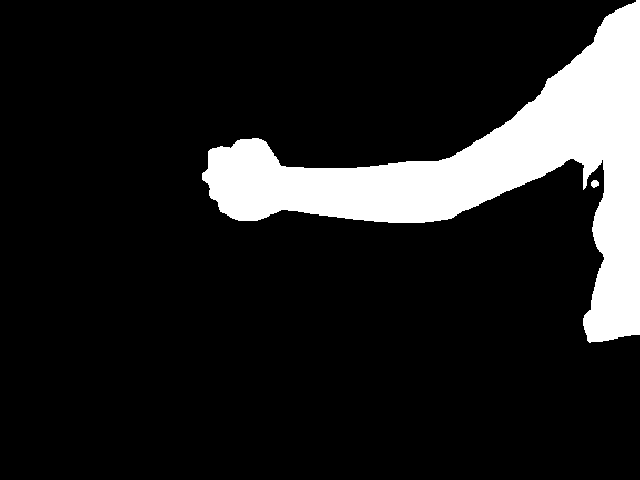}} &
        \centered{\includegraphics[width=8em]{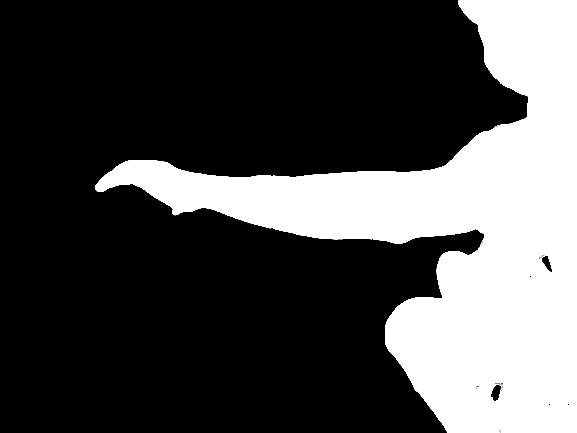}} \\
        \rotatebox[origin=c]{90}{\cite{pixlr2021website}} &
        \centered{\includegraphics[width=8em]{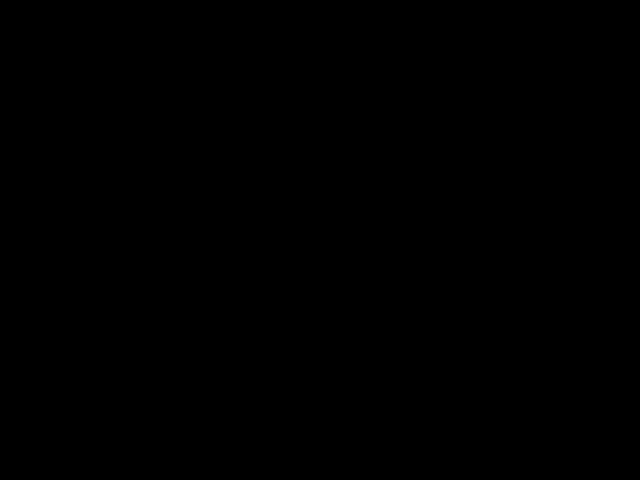}} &
        \centered{\includegraphics[width=8em]{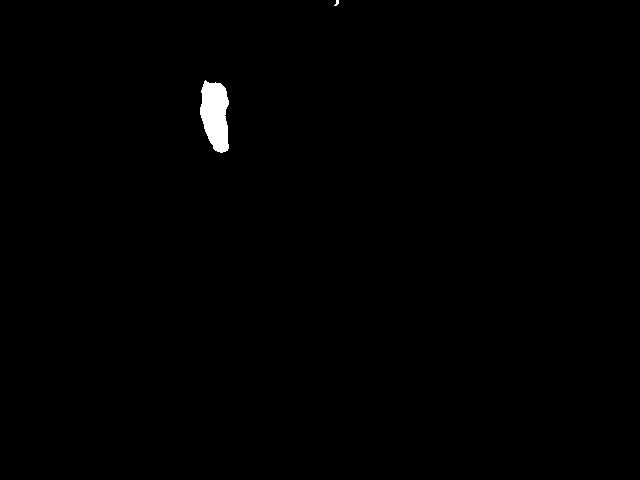}} &
        \centered{\includegraphics[width=8em]{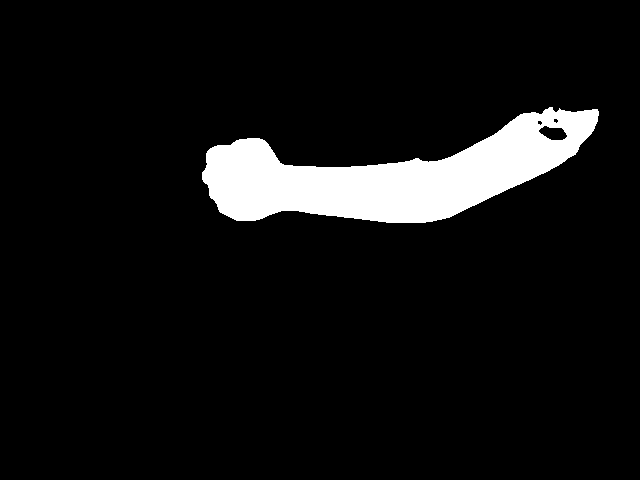}} &
        \centered{\includegraphics[width=8em]{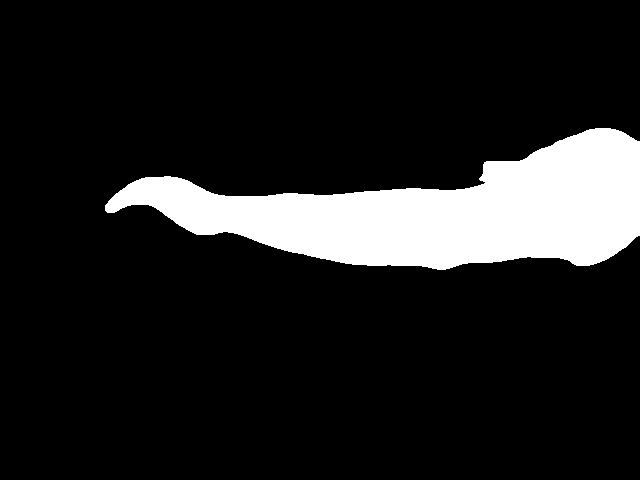}} \\
        \rotatebox[origin=c]{90}{\cite{slazzer2021website}} &
        \centered{\includegraphics[width=8em]{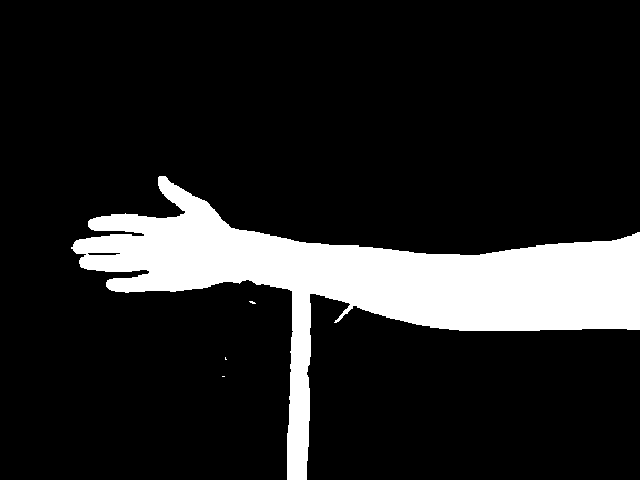}} &
        \centered{\includegraphics[width=8em]{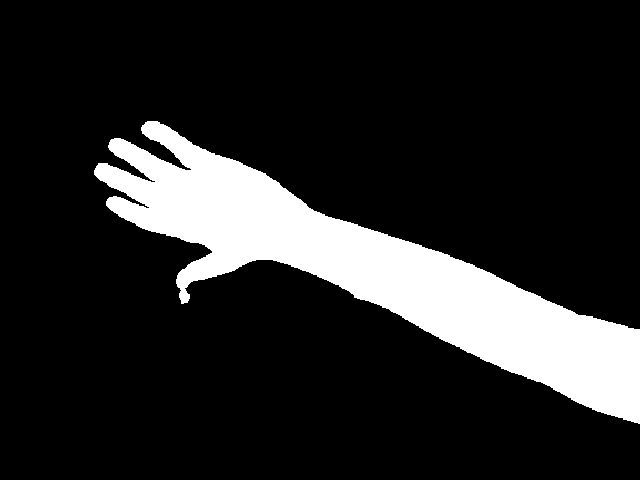}} &
        \centered{\includegraphics[width=8em]{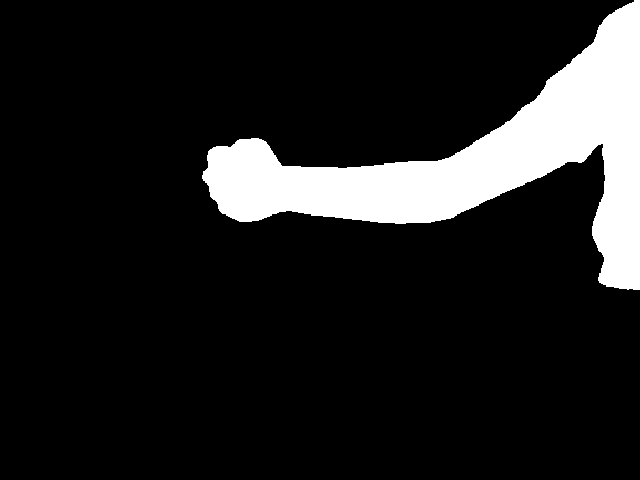}} &
        \centered{\includegraphics[width=8em]{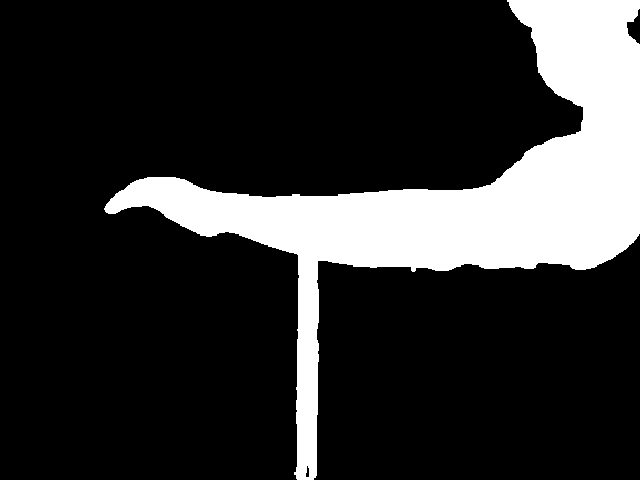}} \\
        \rotatebox[origin=c]{90}{\cite{photoshop2021website}} &
        \centered{\includegraphics[width=8em]{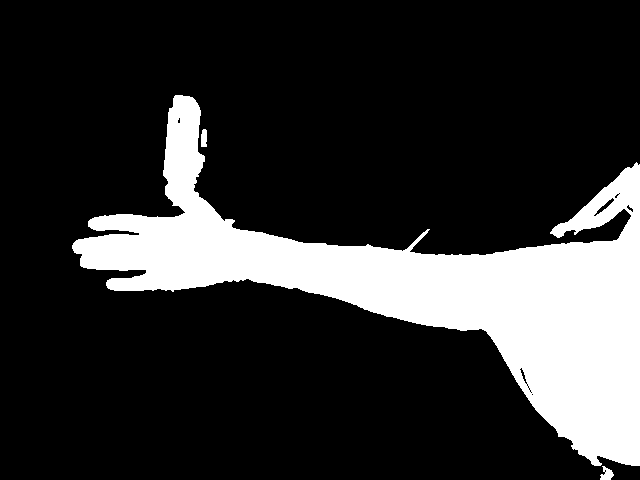}} &
        \centered{\includegraphics[width=8em]{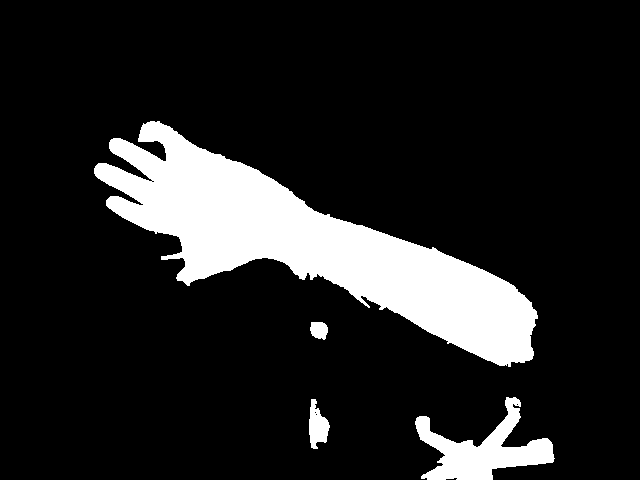}} &
        \centered{\includegraphics[width=8em]{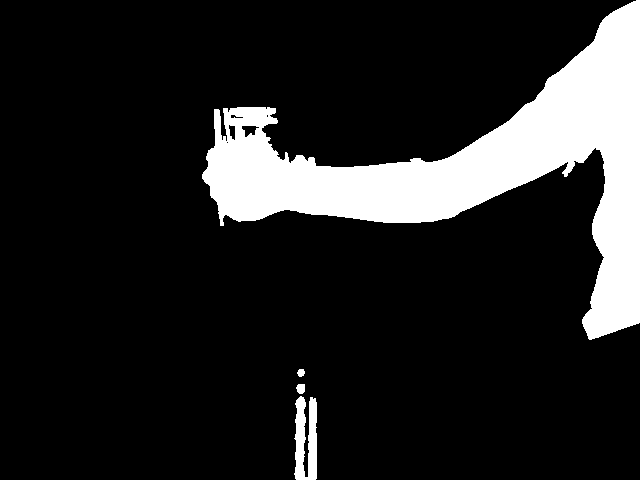}} &
        \centered{\includegraphics[width=8em]{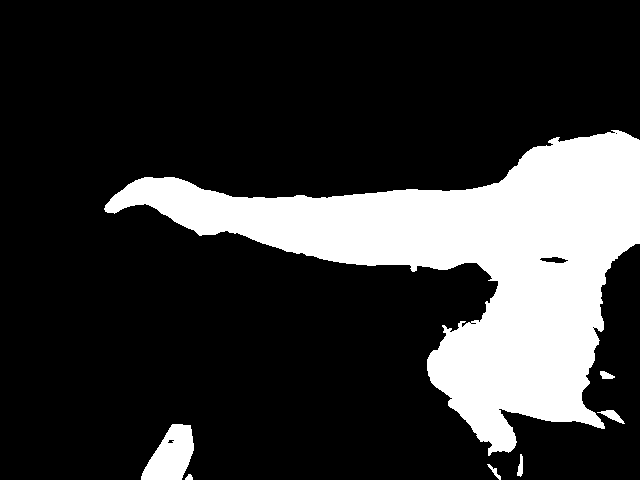}} \\
        \rotatebox[origin=c]{90}{\cite{removalai2021website}} &
        \centered{\includegraphics[width=8em]{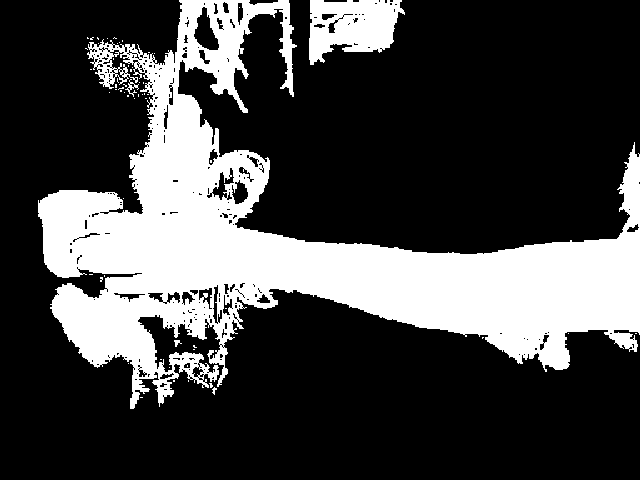}} &
        \centered{\includegraphics[width=8em]{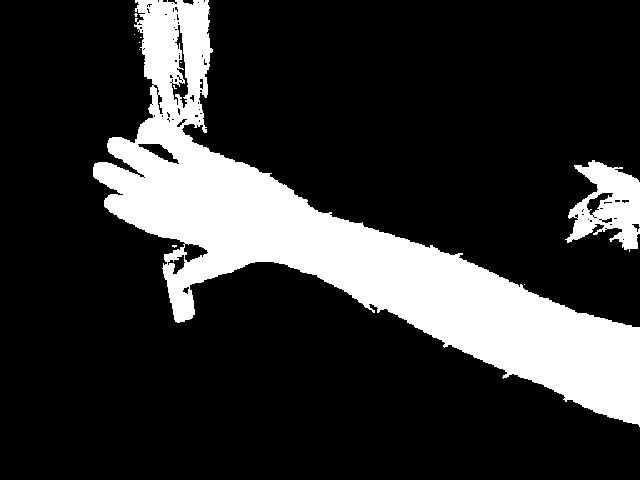}} &
        \centered{\includegraphics[width=8em]{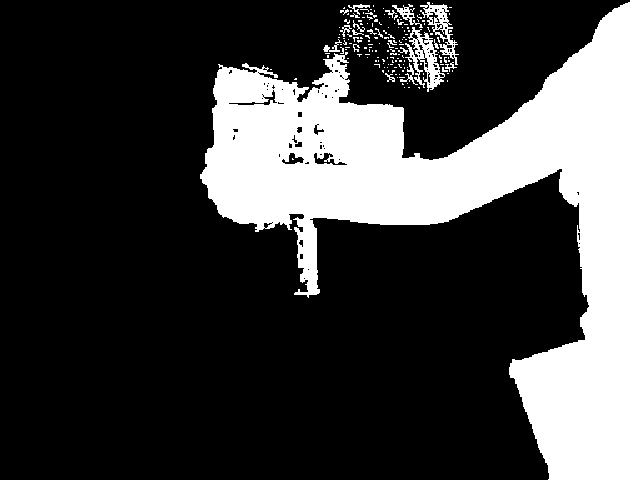}} &
        \centered{\includegraphics[width=8em]{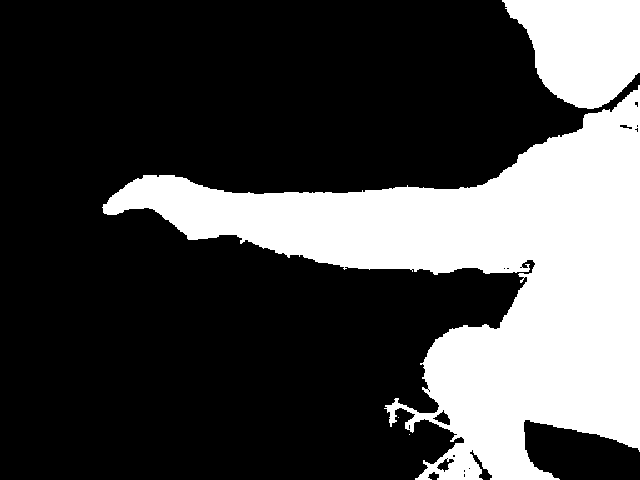}} \\
        \rotatebox[origin=c]{90}{\cite{photoscissors2021website}} &
        \centered{\includegraphics[width=8em]{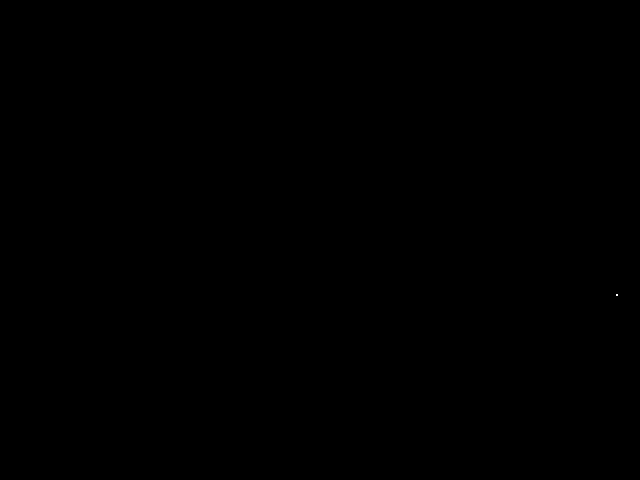}} &
        \centered{\includegraphics[width=8em]{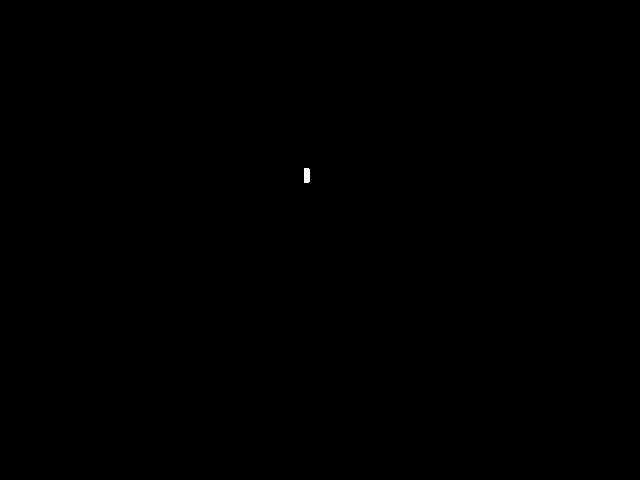}} &
        \centered{\includegraphics[width=8em]{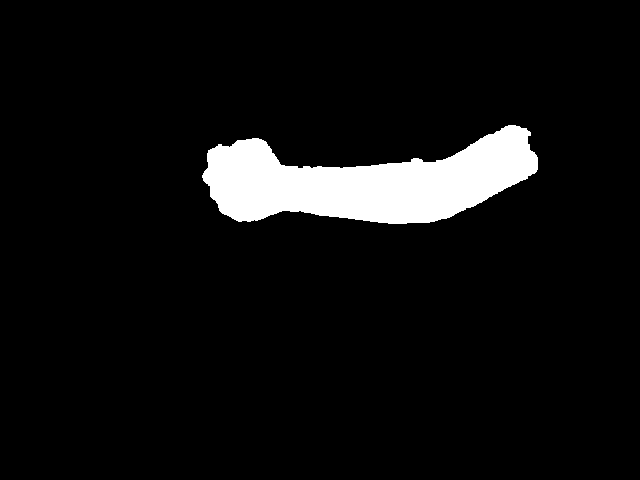}} &
        \centered{\includegraphics[width=8em]{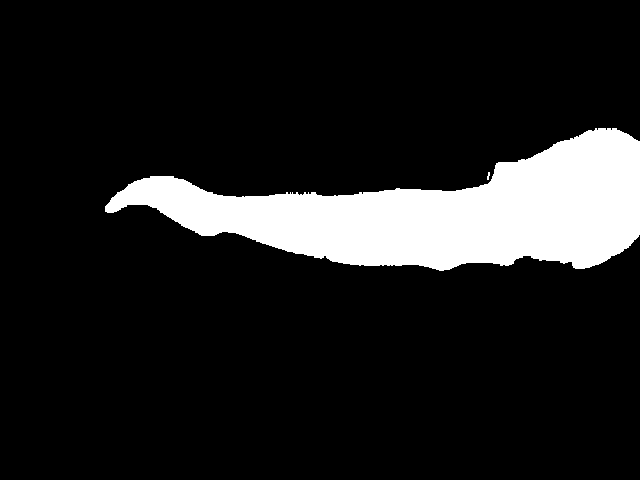}} \\
        \bottomrule
    \end{tabular}
    \caption{Qualitative comparison of the predicted segmentation masks produced by the proposed system and several cutting-edge commercial foreground segmentation models, namely RemoveBG \cite{removebg2021website}, Pixlr \cite{pixlr2021website}, Slazzer \cite{slazzer2021website}, Photoshop \cite{photoshop2021website}, RemovalAI \cite{removalai2021website}, and PhotoScissors \cite{photoscissors2021website}.}
    \label{tab:our_vs_commercial}
\end{table}
\endgroup

\begingroup
\begin{table}
    \centering
    \begin{tabular}{c c c c c c}
        \toprule
        & \multirow{2}{*}{Image} & \multirow{2}{*}{Ground Truth} & \multicolumn{2}{c}{Output} \\
        \cmidrule(lr){4-5}        
        & & & w/ Pre-process & w/out Pre-process \\
        \midrule
        0 &
        \centered{\includegraphics[width=8em]{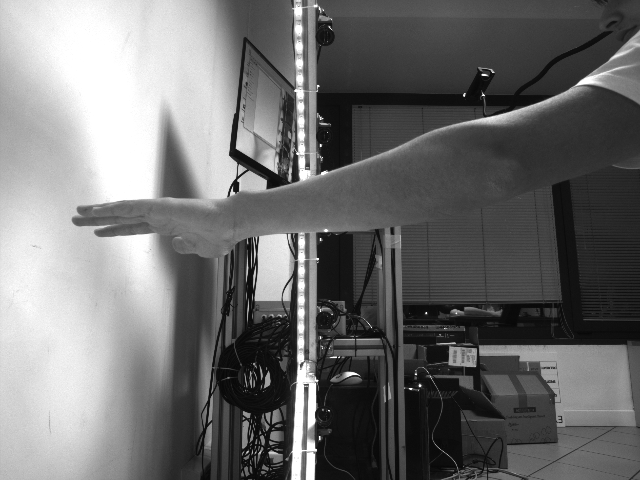}} &
        \centered{\includegraphics[width=8em]{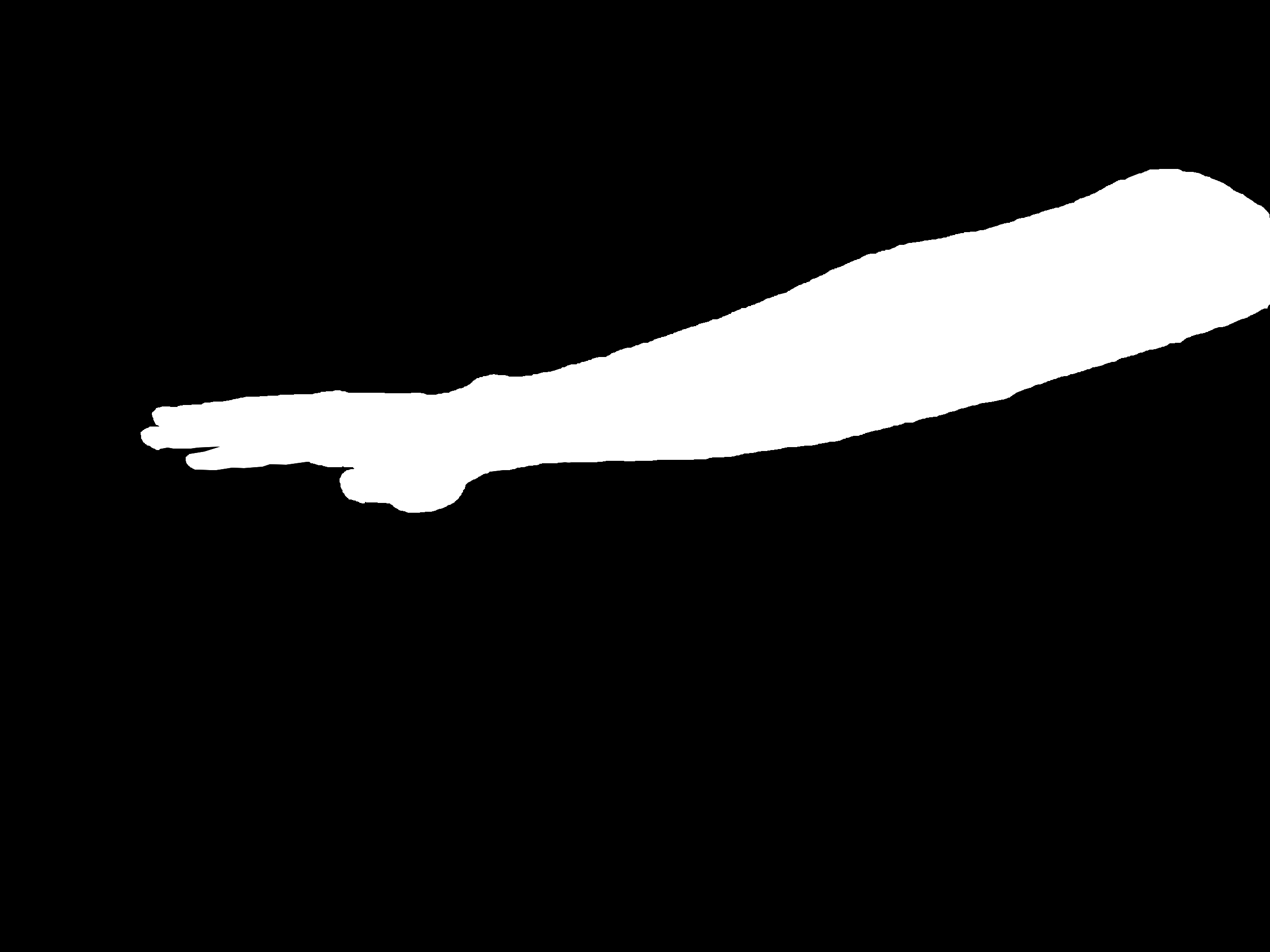}} &
        \centered{\includegraphics[width=8em]{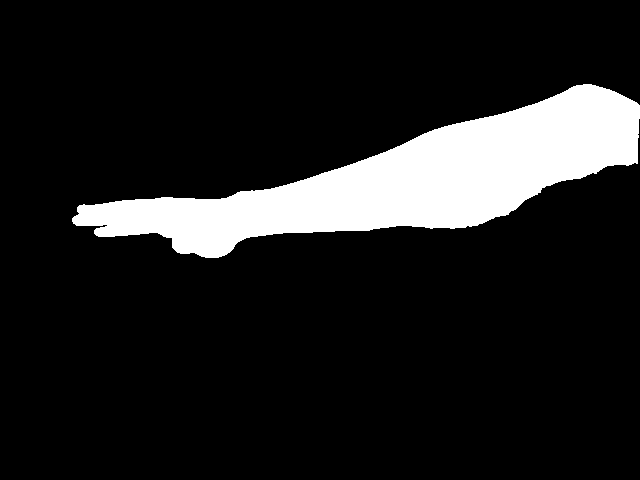}} &
        \centered{\includegraphics[width=8em]{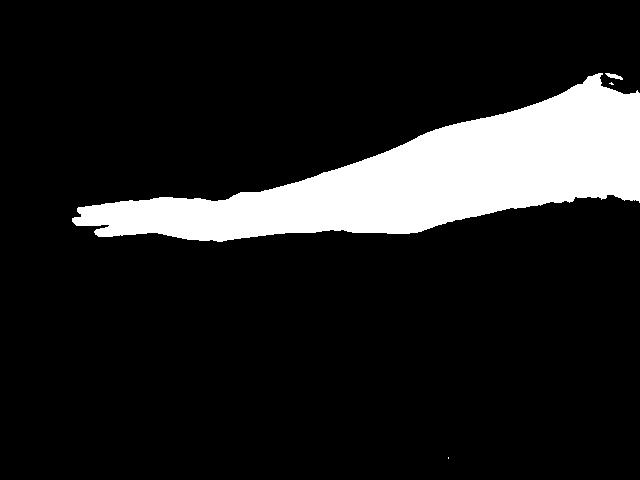}} \\
        1 &
        \centered{\includegraphics[width=8em]{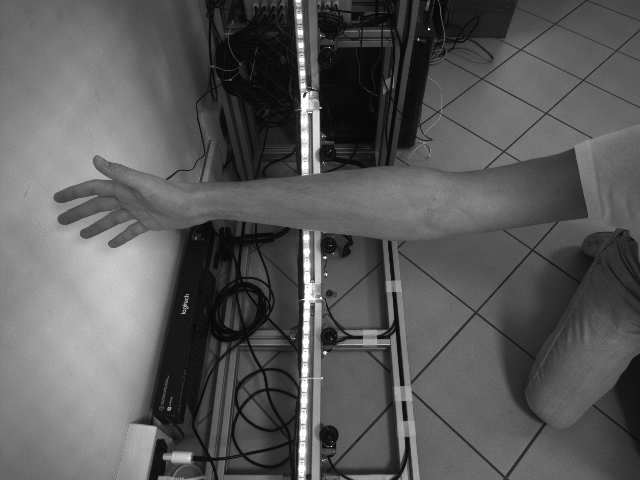}} &
        \centered{\includegraphics[width=8em]{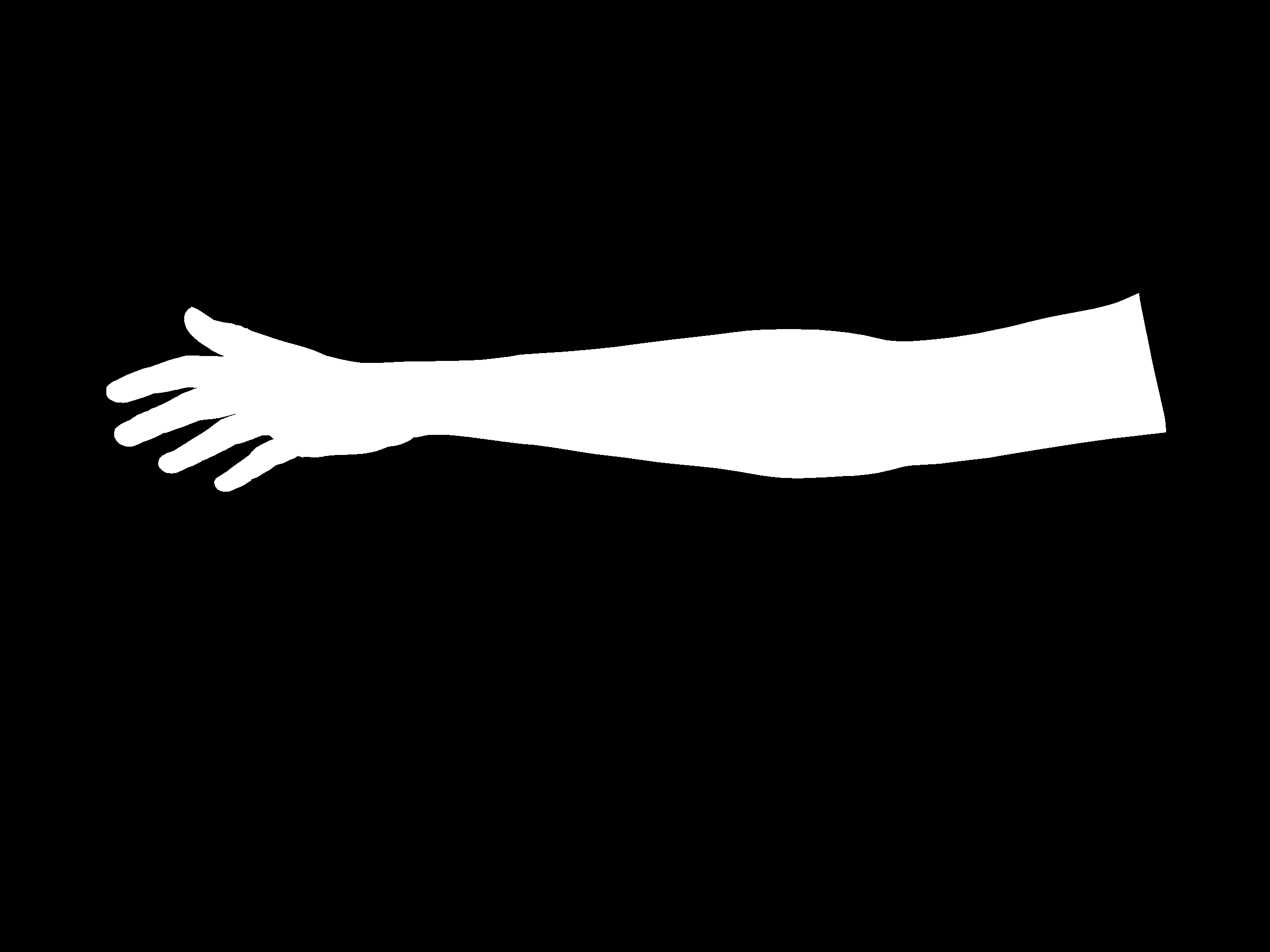}} &
        \centered{\includegraphics[width=8em]{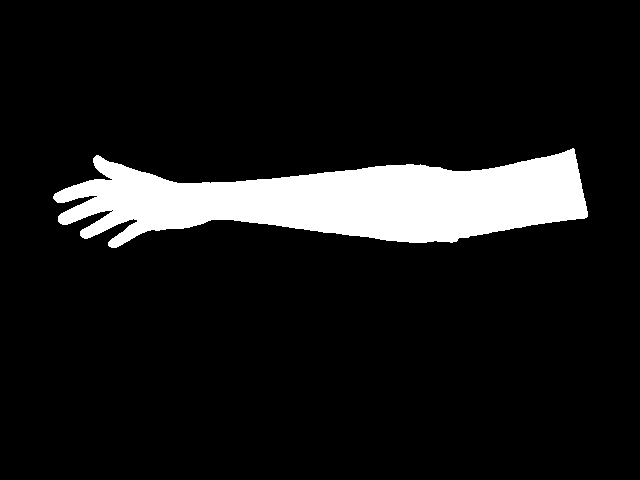}} &
        \centered{\includegraphics[width=8em]{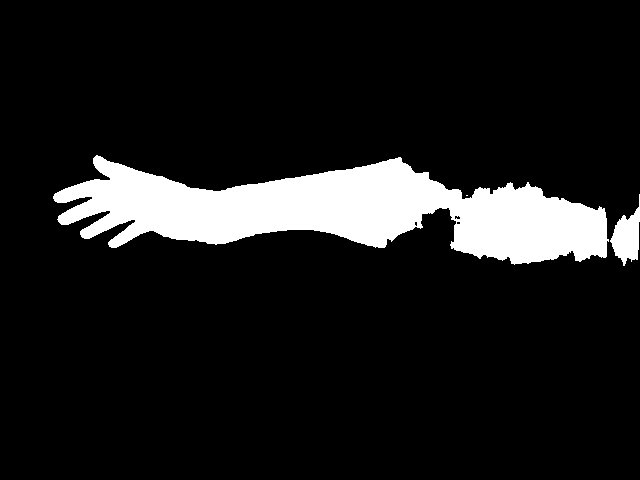}} \\
        2 &
        \centered{\includegraphics[width=8em]{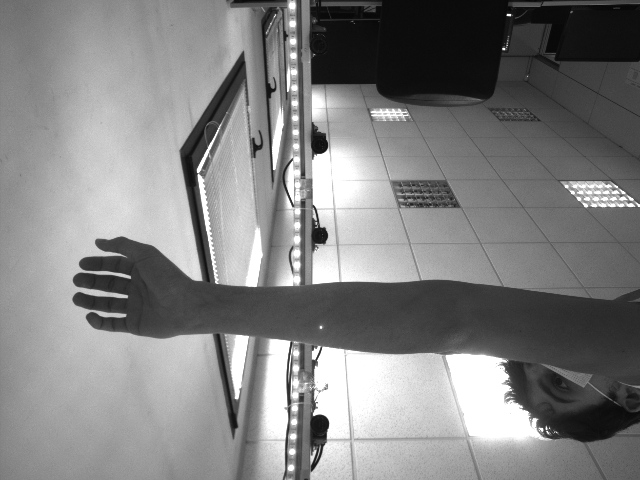}} &
        \centered{\includegraphics[width=8em]{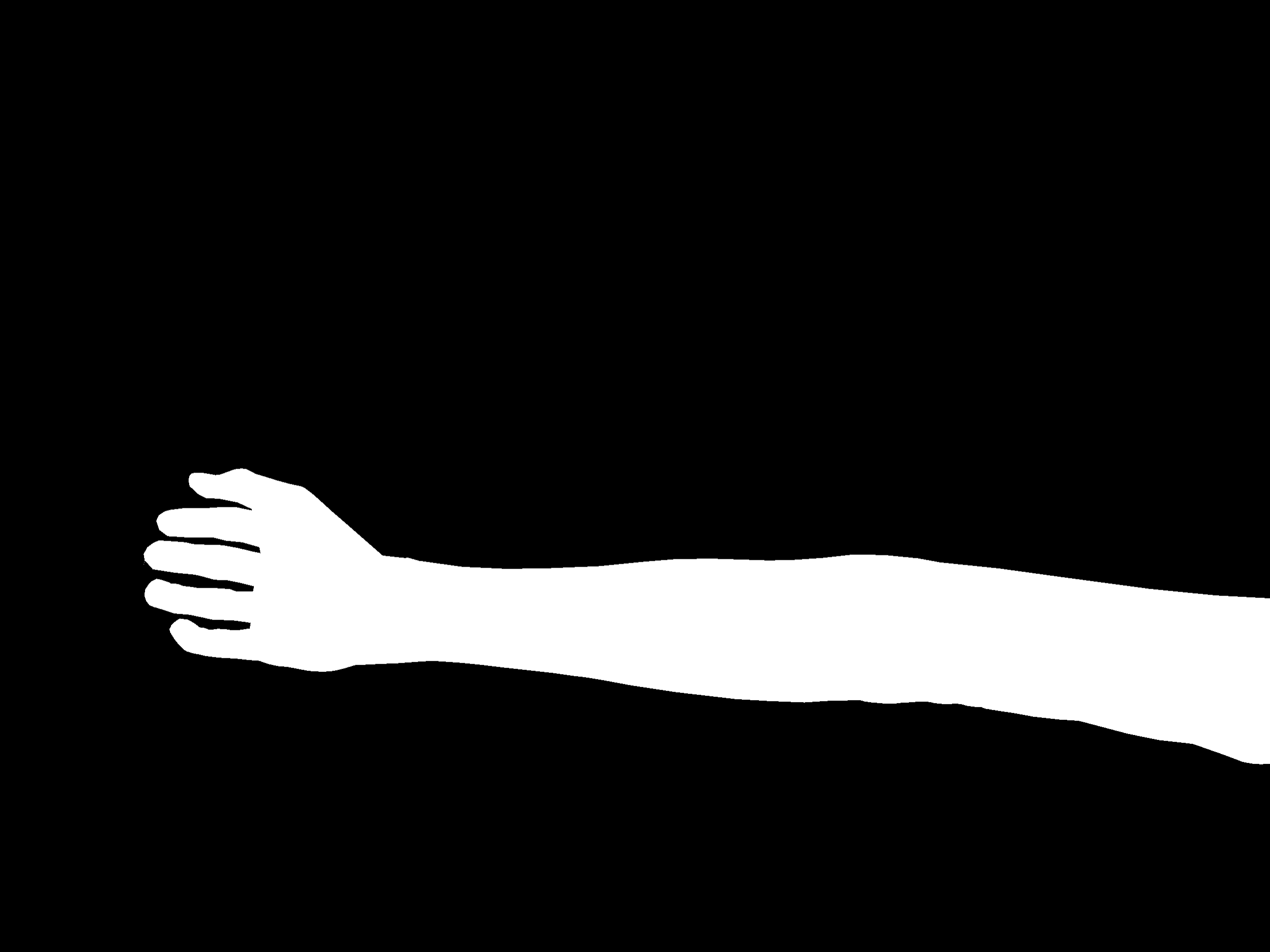}} &
        \centered{\includegraphics[width=8em]{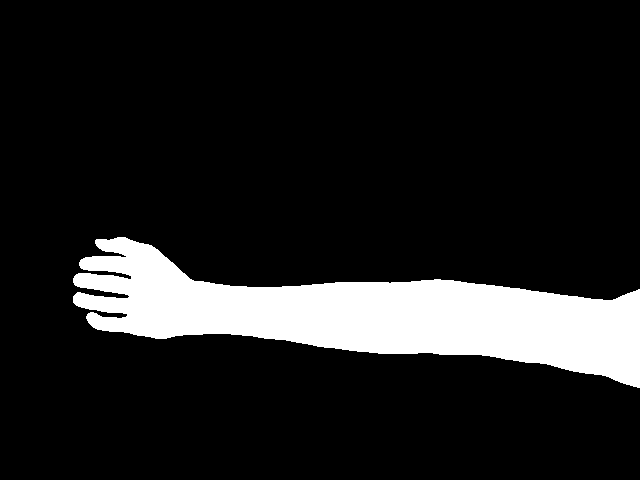}} &
        \centered{\includegraphics[width=8em]{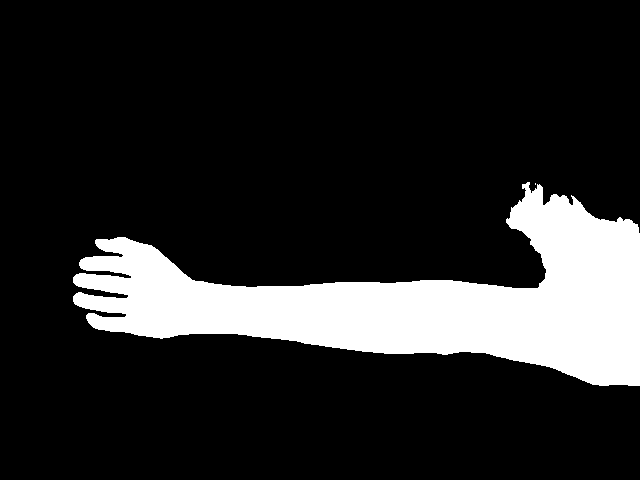}} \\
        \midrule
        3 &
        \centered{\includegraphics[width=8em]{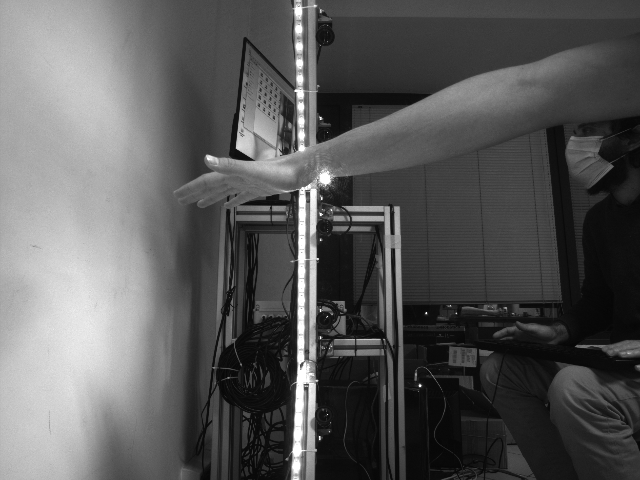}} &
        \centered{\includegraphics[width=8em]{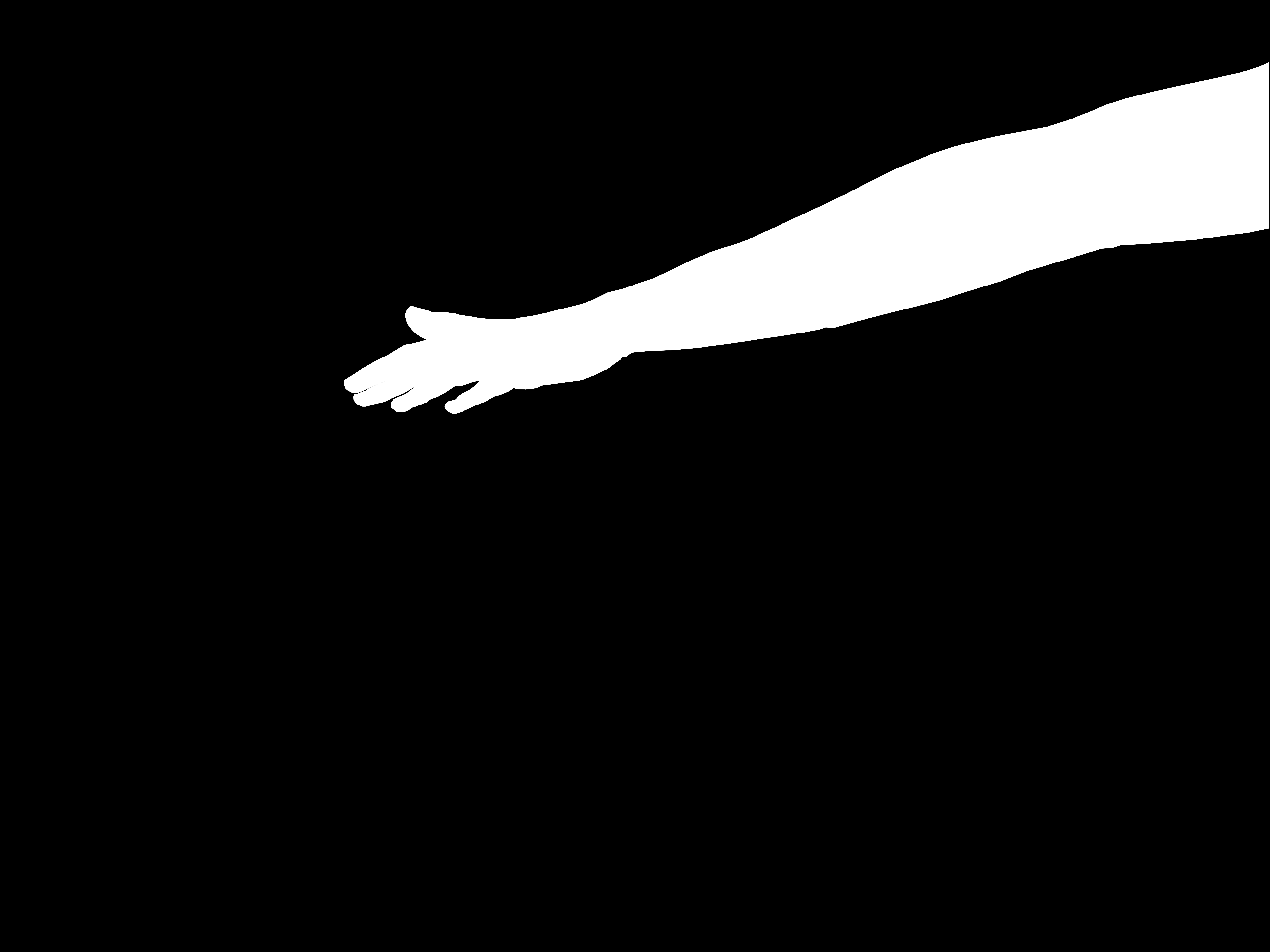}} &
        \centered{\includegraphics[width=8em]{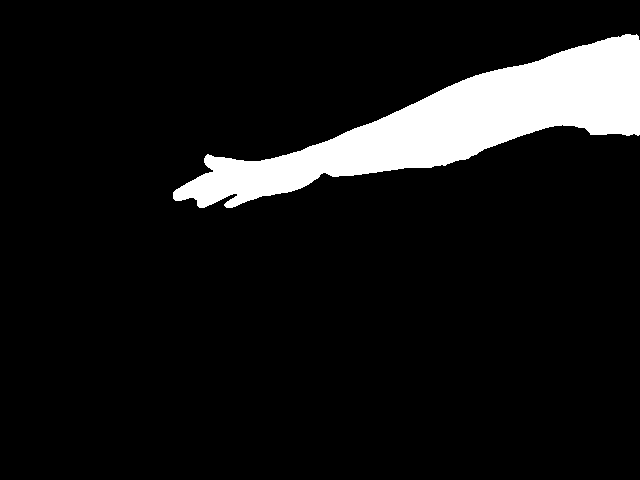}} &
        \centered{\includegraphics[width=8em]{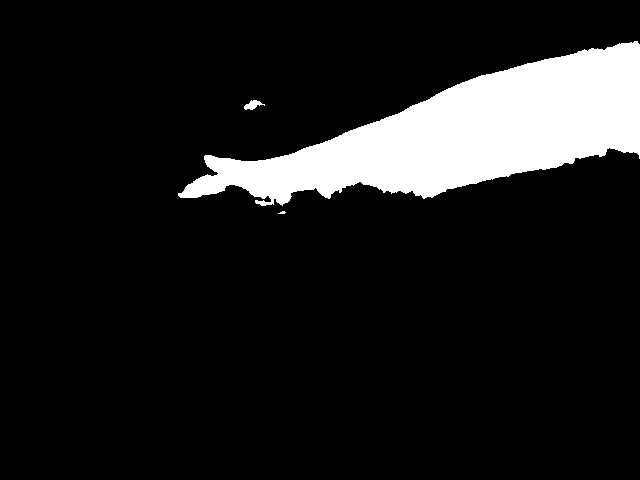}} \\ 
        4 &
        \centered{\includegraphics[width=8em]{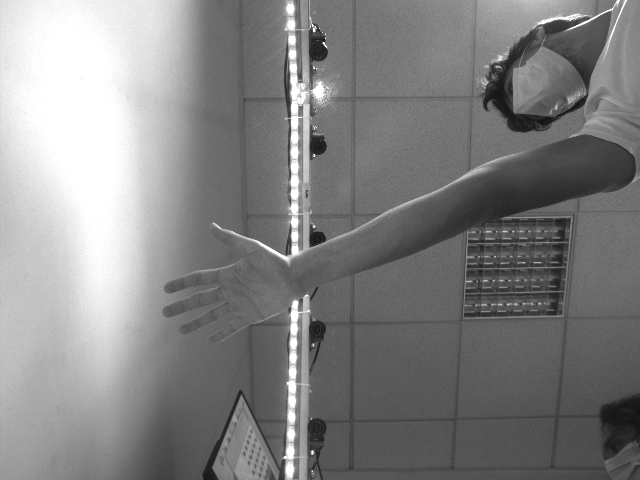}} &
        \centered{\includegraphics[width=8em]{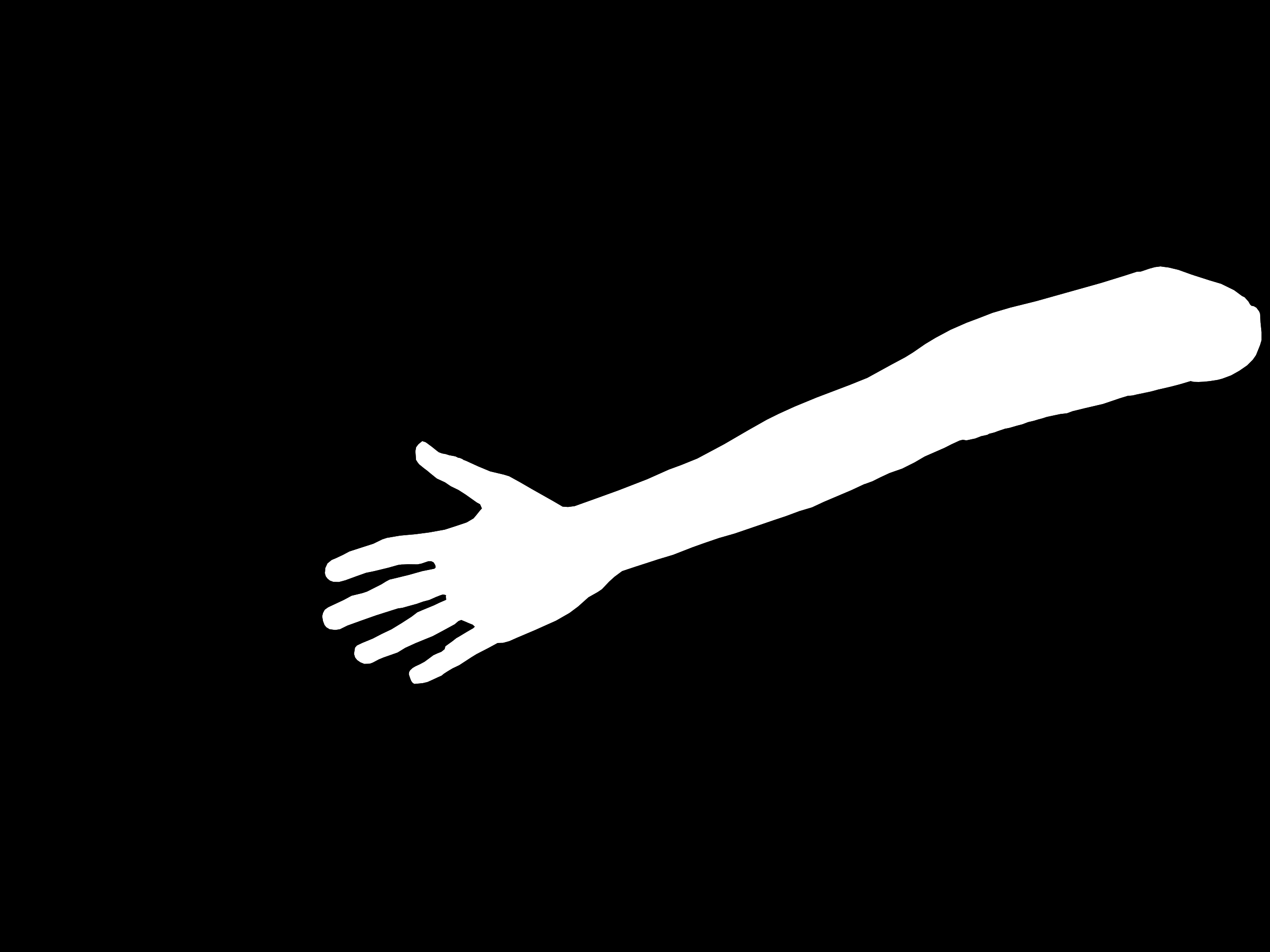}} &
        \centered{\includegraphics[width=8em]{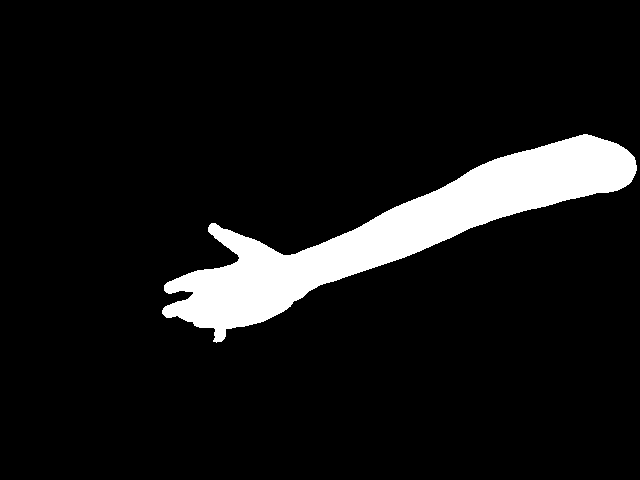}} &
        \centered{\includegraphics[width=8em]{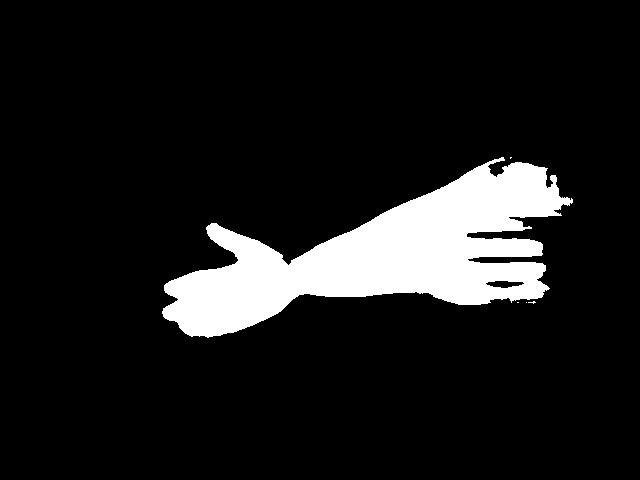}} \\ 
        5 &
        \centered{\includegraphics[width=8em]{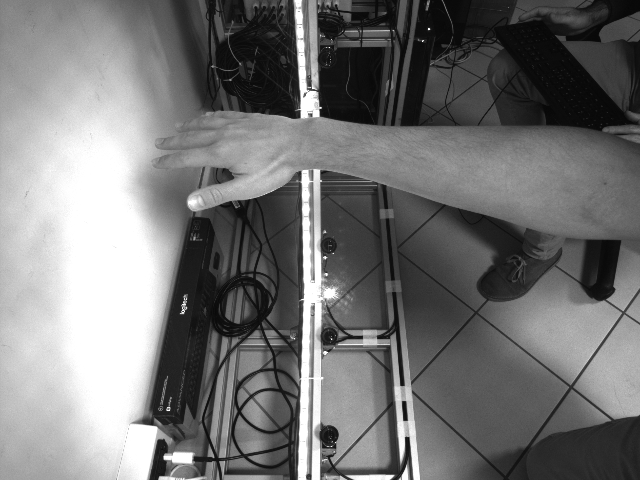}} &
        \centered{\includegraphics[width=8em]{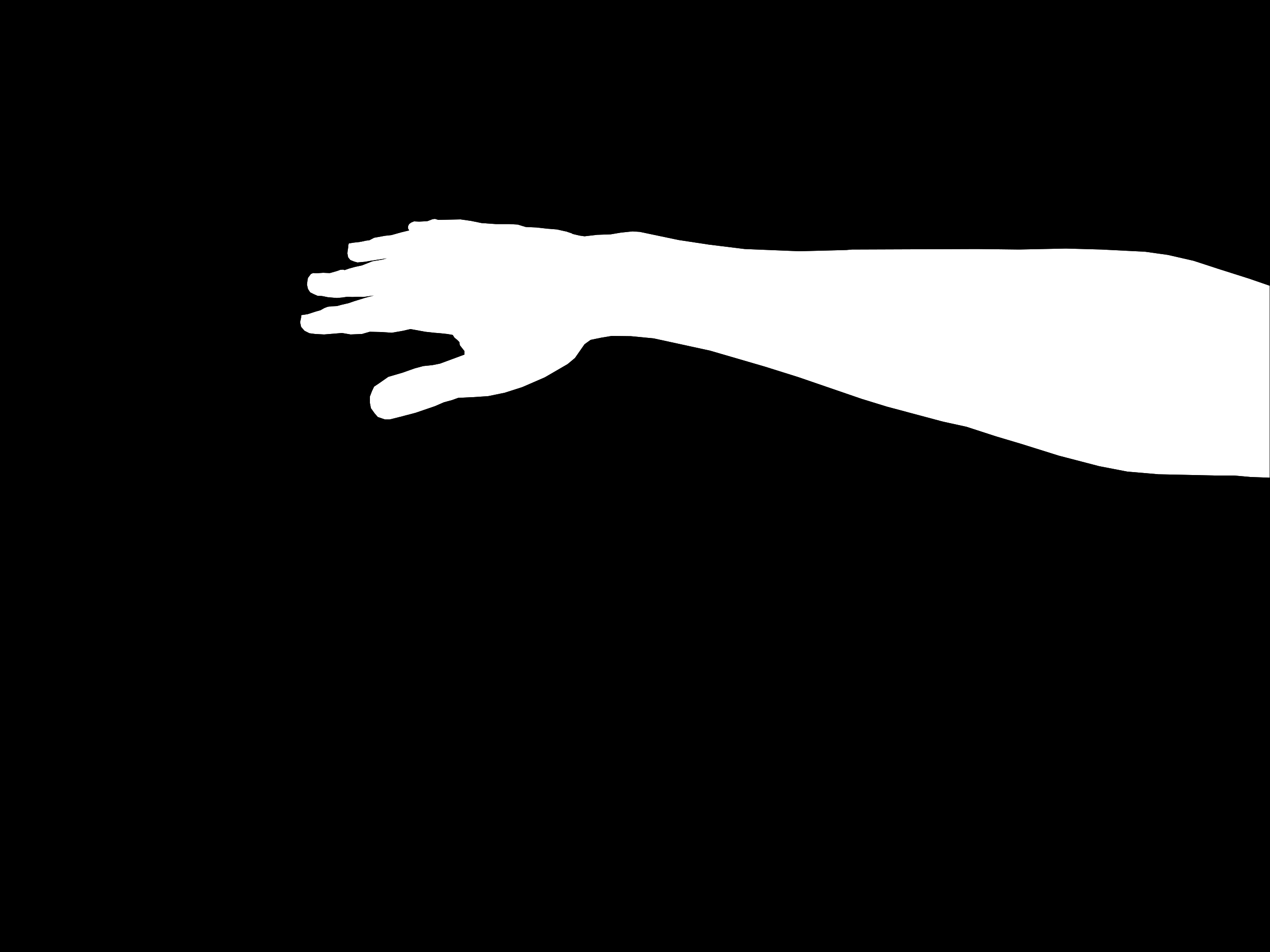}} &
        \centered{\includegraphics[width=8em]{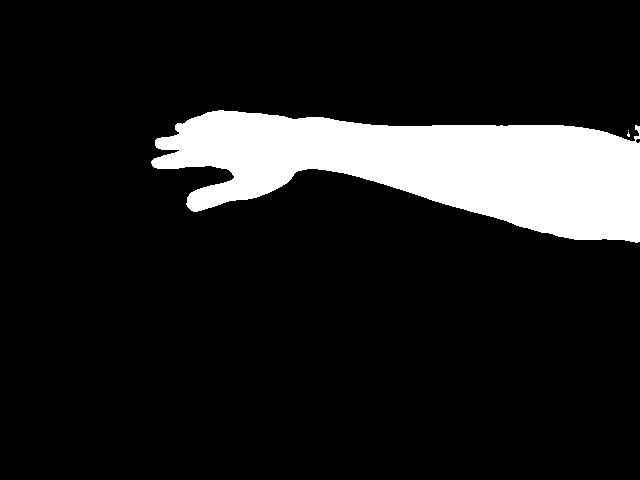}} &
        \centered{\includegraphics[width=8em]{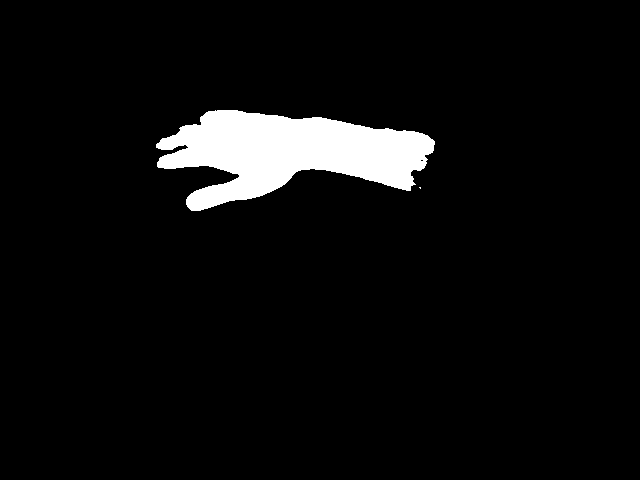}} \\ 
        \midrule
        6 &
        \centered{\includegraphics[width=8em]{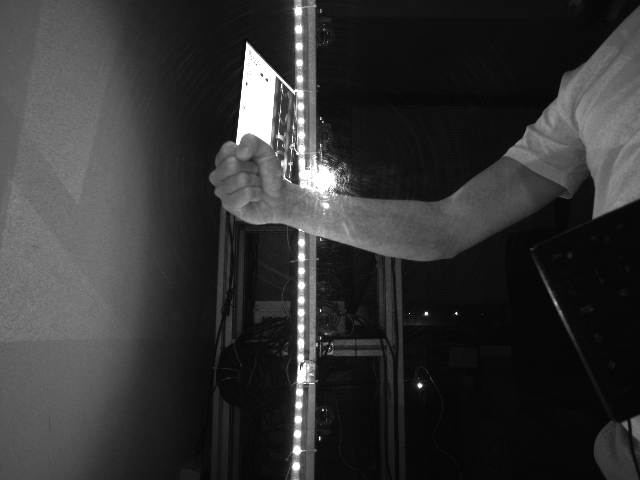}} &
        \centered{\includegraphics[width=8em]{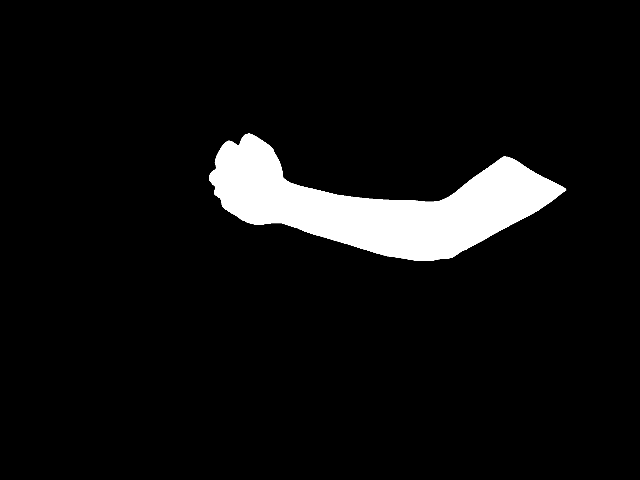}} &
        \centered{\includegraphics[width=8em]{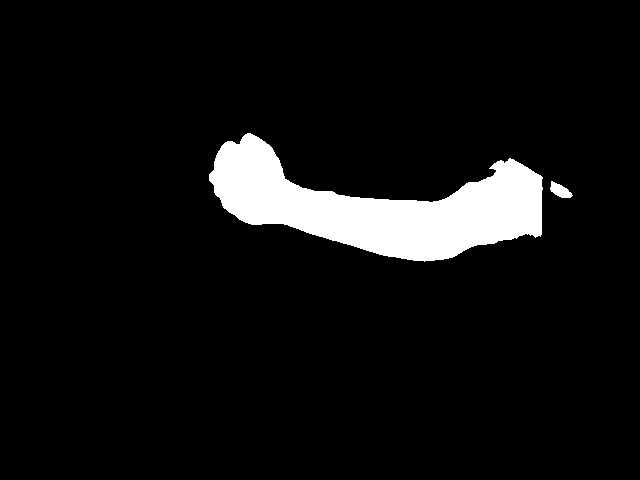}} &
        \centered{\includegraphics[width=8em]{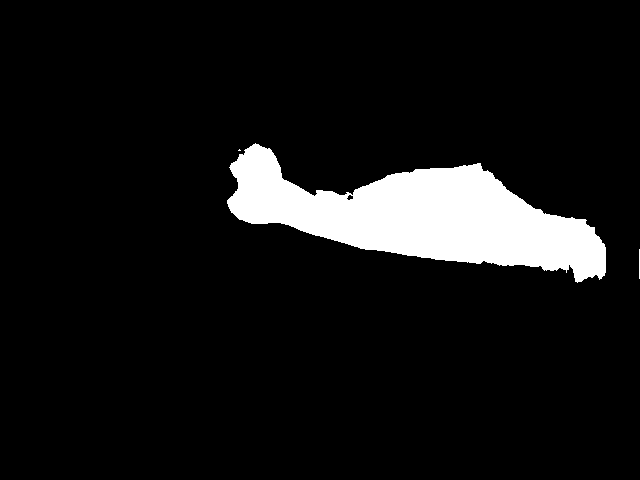}} \\ 
        7 &
        \centered{\includegraphics[width=8em]{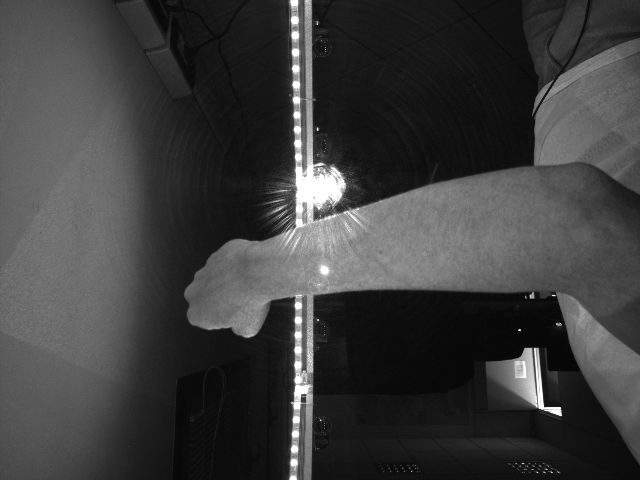}} &
        \centered{\includegraphics[width=8em]{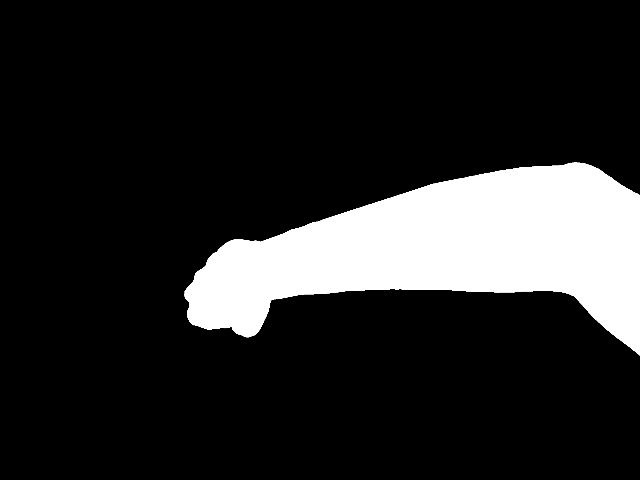}} &
        \centered{\includegraphics[width=8em]{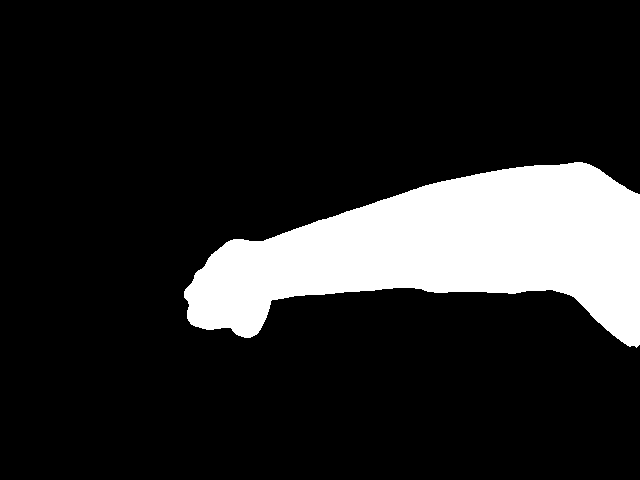}} &
        \centered{\includegraphics[width=8em]{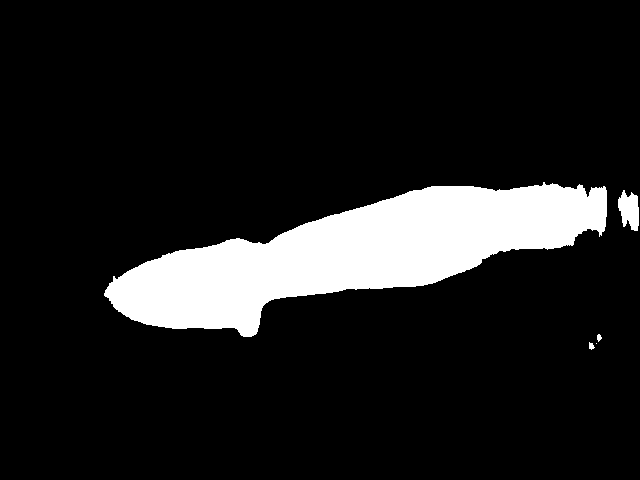}} \\ 
        8 &
        \centered{\includegraphics[width=8em]{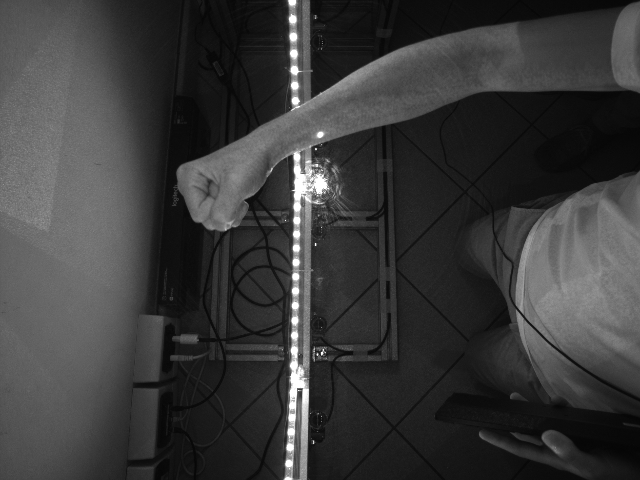}} &
        \centered{\includegraphics[width=8em]{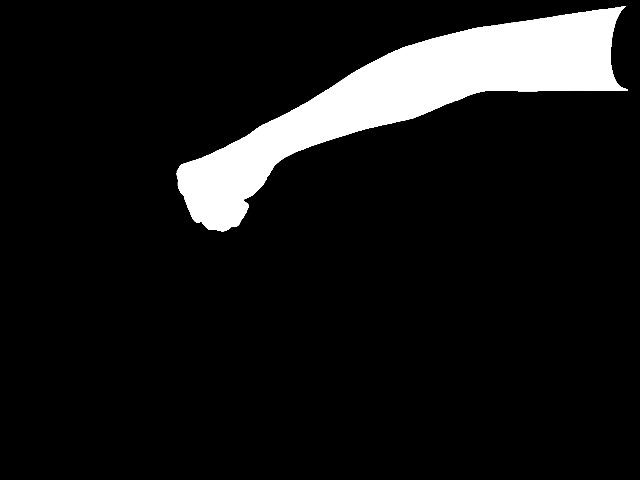}} &
        \centered{\includegraphics[width=8em]{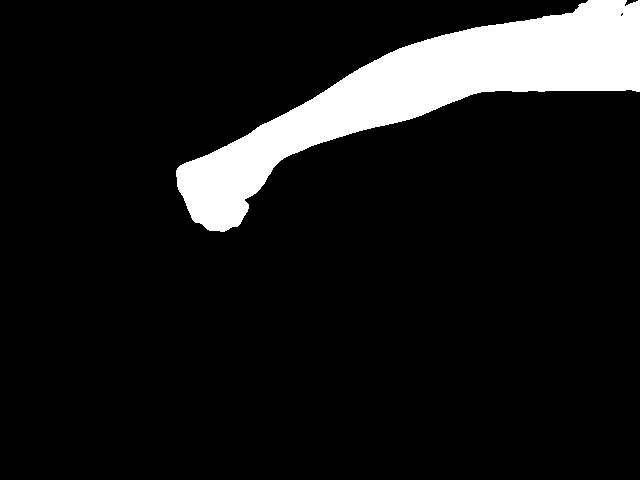}} &
        \centered{\includegraphics[width=8em]{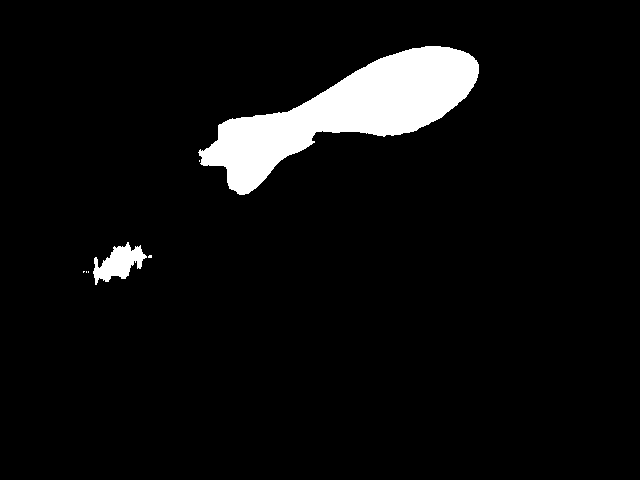}} \\ 
        \bottomrule
    \end{tabular}
    \caption{Qualitative comparison of the predicted segmentation masks produced by the proposed system with and without the pre-processing module. Images $0,1,2$ are taken from the High test set, images $3,4,5$ are taken from the Medium test set, and images $6,7,8$ are taken from the Low test set.}
    \label{tab:preprocess_ablation_qual}
\end{table}
\endgroup

\clearpage

\subsection{Ablation Study}
\label{subsec:ablation_study}

\begingroup
\setlength{\tabcolsep}{5pt}
\begin{table}[t]
    \centering
    \begin{tabular}{c c c c c c}
        \toprule
        \multicolumn{3}{c}{Input Tensor} & \multirow{2}{*}{High} & \multirow{2}{*}{Medium} & \multirow{2}{*}{Low}\\
        \cmidrule(lr){1-3}
        ch1 & ch2 & ch3 & & &\\
        \midrule
        RAW & RAW & RAW & 0.808 & 0.734 & 0.499 \\
        HED & HED & HED & 0.922 & 0.951 & 0.849 \\
        HED & RAW & RAW & 0.936 & 0.951 & 0.836 \\
        HED & CLAHE2 & CLAHE4 & \textbf{0.946} & \textbf{0.959} & \textbf{0.892} \\
        \bottomrule
    \end{tabular}
    \caption{Ablation study of the pre-processing module on the High, Medium and Low test sets. Segmentation scores are reported in terms of the intersection on the union. For each test set, the best score is reported in \textbf{bold}.}
    \label{tab:ablation_quantitative}
\end{table}
\endgroup

\begin{figure}[t]
    \centering
    \includegraphics[width=0.85\linewidth]{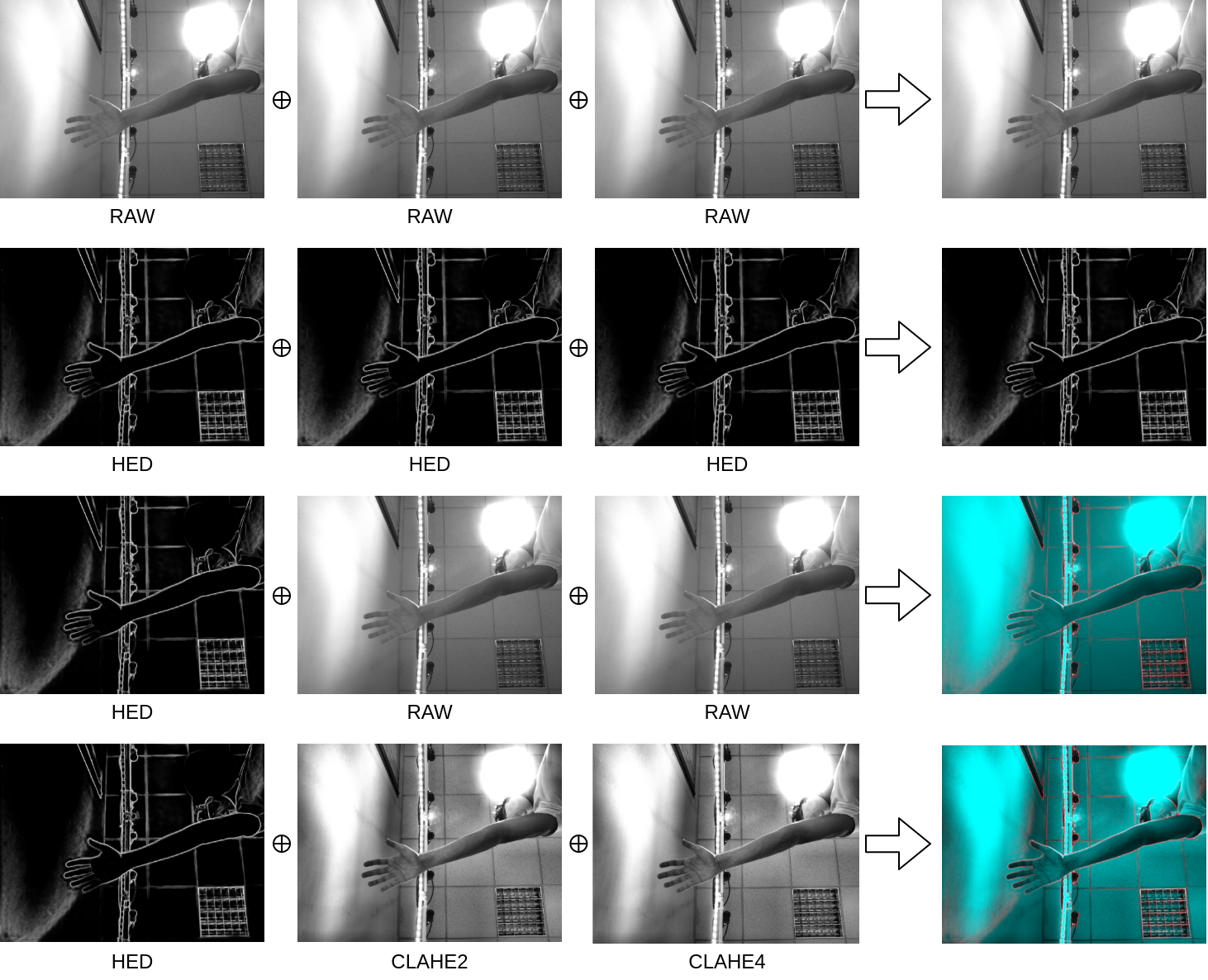}
    \vspace{-0.5em}
    \caption{Visual comparison of the ablation study conducted in Sec.~\ref{subsec:ablation_study}.
    In each row, three images are concatenated to form the input to the segmentation network. The last row illustrates the transformation applied by the pre-processing module.
    }
    \label{fig:ablation_qualitative}
\end{figure}

To study the impact of the single transformations applied by the pre-processing module to the input of the segmentation network, we produced an ablation study by systematically omitting such transformations from the input tensor. As introduced in Sec.~\ref{sec:network}, the pre-processing module generates a tensor of three transformed images. For each test of this study, we omit such transformations but for ease of interpretability and comparison, we keep the size of the tensor to three images by repeating the raw input. Moreover, for this study, we used the UNet++ segmentation network with a SE-ResNet-50 encoder pre-trained on ImageNet. We trained the network on the proposed synthetic dataset by systematically changing the pre-processing transformations. Therefore, in each study, the segmentation network was trained on different data. As in the previous sections, we trained the network for $66$K iterations using a batch size of 4 and we used Adam optimizer with a learning rate and epsilon equal to $0.0001$.

Tab.~\ref{tab:ablation_quantitative} lists the results of omitting individual transformations from the tensor given in input to the UNet++ network. Each row of the table presents the IoU score obtained by evaluating on the High, Medium and Low test sets a UNet++ network trained on a different transformed synthetic dataset. The first row shows the results obtained by the segmentation network when the pre-processing module is omitted. This case was presented both qualitatively and quantitatively in Sec.~\ref{subsec:performance_evaluation}. The IoU score drops by up to $45$\%, thus showing the importance of pre-processing the input data.
The second row lists the results obtained by the segmentation network when the input data is preprocessed by only using the holistically-nested edge detection (HED). Compared with the RAW data input test, it provides a sensible improvement in terms of IoU in all test sets. However, we noted that the exclusive use of HED determines a great loss of information, for example, we no longer consider their texture. As a consequence, we tried to combine the edge response map with the original raw image. As shown in the third row of Table~\ref{tab:ablation_quantitative}, this solution achieves a higher score on the High test set, but it leads to a slight performance deterioration on the Low test set. This demonstrates the poor generalization capabilities for varying illumination conditions of the segmentation network trained using a mixture of raw and pre-preprocessed data. To improve robustness to large light conditions changes, we finally included image equalization in the pre-processing module. The last row of the table presents the results obtained by the segmentation network when using our final pre-processing module, i.e., the input data is pre-processed by using HED and CLAHE. Fig.~\ref{fig:ablation_qualitative} provides a visual comparison of the input tensors analyzed by this ablation study.

\section{Conclusions}

In this paper, we proposed a novel framework for human body part segmentation, based on state-of-the-art Deep Convolutional Neural Networks, that does not require manually annotated data. Training data is automatically generated by rendering with a game engine synthetic, but photo-realistic, views of human limbs. Random geometric non-rigid transformations are applied to the rendered models as well as different illumination conditions. In this way, the proposed data generation pipeline can generate multiple unique synthetic images and their corresponding annotations.
A pre-processing module that combines an edge response map with equalized versions of the input image has been specifically designed to polarize the network to focus on the shape of the objects of interest (human limbs in our case) rather than the skin tone (intensity or color) or its characteristic texture. Exhaustive performance evaluations on a custom-built, manually labeled dataset that includes several views of real human arms show the effectiveness of our method. An open-source implementation of our system along with the produced datasets is made publicly available with this paper.

\section*{Acknowledgment}

This project has received funding from the DIGI-B-CUBE voucher framework which has been supported by the European Union’s Horizon 2020 research and innovation program under grant agreement No 824920.
Part of this work has been supported by MIUR (Italian Minister for Education) under the initiative “Departments of Excellence” (Law 232/2016).


\bibliographystyle{unsrt}
\bibliography{references}

\end{document}